\documentclass[10pt,twocolumn,letterpaper]{article}

\usepackage{iccv}
\usepackage{times}
\usepackage{epsfig}
\usepackage{graphicx}
\usepackage{amsmath}
\usepackage{amssymb}
\usepackage{subfig}
\usepackage{booktabs}
\usepackage{multirow}
\usepackage{mathtools}
\usepackage{comment}
\usepackage{adjustbox}
\usepackage{wrapfig}
\usepackage{times}
\usepackage{dsfont}
\usepackage{epsfig}
\usepackage{microtype}
\usepackage{cite}
\usepackage{enumitem}
\usepackage{tabularx} 
\usepackage{ mathrsfs }
\usepackage{color, colortbl}

\definecolor{Gray}{gray}{0.9}
\usepackage[toc,page]{appendix}

\usepackage[pagebackref=true,breaklinks=true,letterpaper=true,colorlinks,bookmarks=false]{hyperref}

\iccvfinalcopy 

\def\iccvPaperID{12566} 

\ificcvfinal\fi

\newcommand*{\affmark}[1][*]{\textsuperscript{#1}}
\usepackage[accsupp]{axessibility}

\begin{document}

\title{MEGA: Multimodal Alignment Aggregation and Distillation\\ For Cinematic Video Segmentation}

\author{Najmeh Sadoughi\affmark[1], Xinyu Li\affmark[1], Avijit Vajpayee\affmark[1], David Fan\affmark[1], Bing Shuai\affmark[2], \\Hector Santos-Villalobos\affmark[1], Vimal Bhat\affmark[1], Rohith MV\affmark[1]\\
\affmark[1]Amazon Prime Video, \affmark[2]AWS AI Labs\\
{\tt\small \{nnnourab,xxnl,avivaj,fandavi,bshuai,hsantosv,vimalb,kurohith\}@amazon.com}
}

\maketitle
\ificcvfinal\thispagestyle{empty}\fi


\begin{abstract}
    \noindent Previous research has studied the task of segmenting cinematic videos into scenes and into narrative acts. However, these studies have overlooked the essential task of multimodal alignment and fusion for effectively and efficiently processing long-form videos ($>60$min). In this paper, we introduce \textbf{M}ultimodal alignm\textbf{E}nt a\textbf{G}gregation and distill\textbf{A}tion (MEGA) for cinematic long-video segmentation. MEGA tackles the challenge by leveraging multiple media modalities. The method coarsely aligns inputs of variable lengths and different modalities with alignment positional encoding. To maintain temporal synchronization while reducing computation, we further introduce an enhanced bottleneck fusion layer which uses temporal alignment. Additionally, MEGA employs a novel contrastive loss to synchronize and transfer labels across modalities, enabling act segmentation from labeled synopsis sentences on video shots. Our experimental results show that MEGA outperforms state-of-the-art methods on MovieNet dataset for scene segmentation (with an Average Precision improvement of +1.19\%) and on TRIPOD dataset for act segmentation (with a Total Agreement improvement of +5.51\%). 
\end{abstract}


\section{Introduction} 
\label{sec:intro}
\begin{figure}[t]
    \centering
    \includegraphics[width=0.45\textwidth]{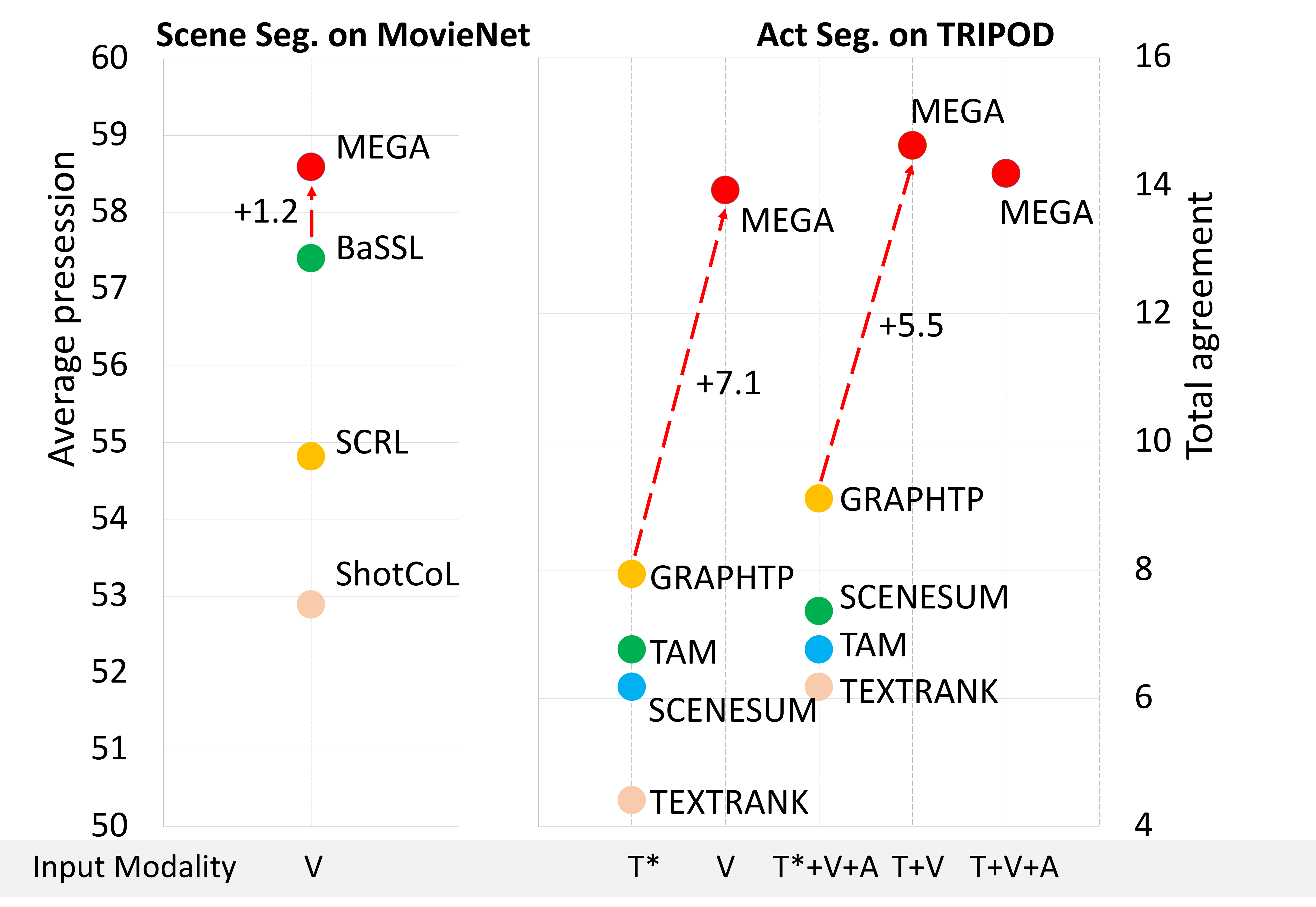}
    \caption{MEGA works well on both scene segmentation and act segmentation tasks, outperforming previous work with significant margin. V,T*,T,A denotes video, screenplay, subtitle and audio respectively.}
    \label{fig:teaser}
    \vspace{-5mm}
\end{figure}

In the world of video production, movies are composed of smaller units called shots, scenes, and acts. Shots are a continuous set of frames, a scene is a sequence of shots that tell a story, and an act is a thematic section of a narrative~\cite{hauge2017storytelling}. 
While computer vision has made significant strides in shot detection~\cite{souvcek2020transnet}, scene and act segmentation remain a challenge, despite their potential for smart video navigation, advertisement insertion, and movie summarization. 
Cinematic content comprises of different data sources, including audio, visual, and text data, as well as derivative data sources from the narrative, including location, appearance, tone, or acoustic events. 
In this work, we will refer to all of these input components as ``modalities" of cinematic content.
Previous work has not fully explored how to align and aggregate these modalities which have different granularities.

We propose to address scene and act segmentation tasks with an unified multimodal Transformer. However, this approach presents two main challenges. 
Firstly, there is the issue of cross modality information synchronization and fusion at the shot level. 
Previous studies which use multimodal fusion for scene and act segmentation perform early~\cite{chen2021shot,papalampidi2021movie} or late fusion of features~\cite{rao2020local}, and have not explored fusion strategies which utilize multimodal temporal alignment. 
Additionally, the fusion strategies that utilize temporal alignment such as merged attention or cross modality attention~\cite{dou2022empirical,kim2021vilt} are computationally expensive and not generalizable to a large number of modalities. 
Secondly, due to the challenges associated with labeling a long video on act segmentation, the labels for act segmentation are provided on synopsis sentences~\cite{papalampidi2019movie} which do not provide timestamps.  
To avoid the more challenging task of cross-modal synchronization, previous studies on act segmentation~\cite{papalampidi2021movie,papalampidi2019movie} rely on textual screenplay to transfer the labels from synopsis to movie, ignoring the rich multimodal information from the video, and introducing an additional dependency on screenplay data which is not always available.

%
To address these challenges, we introduce \textbf{M}ultimodal alignm\textbf{E}nt a\textbf{G}gregation and distill\textbf{A}tion (MEGA). 
MEGA includes a novel module called \textit{alignment positional encoding} which aligns inputs of variable lengths and different modalities at a coarse temporal level. 
To fuse the aligned embeddings of different modalities in an efficient and effective manner, we adopt the bottleneck fusion tokens~\cite{nagrani2021attention} and append a set of fusion tokens to each modality. These tokens share the same sequence length as the normalized positional encoding for different modalities, allowing us to inject them with the coarse temporal information, enabling information fusion in a better aligned embedding space. 
To address the issue of cross-domain knowledge transfer, we introduce a cross-modal synchronization approach. This method allows us to transfer the manually labeled act boundaries from synopsis level to movie level using rich multimodal information, enabling us to train MEGA directly on videos without relying on screenplay -- which was a hard requirement for previous works~\cite{papalampidi2021movie,papalampidi2019movie}.


We test our proposed alignment and aggregation modules on the Movienet-318 ~\cite{huang2020movienet} and the TRIPOD datasets~\cite{papalampidi2019movie}, and we test our cross modality synchronization module on TRIPOD alone, as the labels are provided on a different modality during training.
Our proposed MEGA outperforms previous SoTA on scene segmentation on the Movienet-318 dataset (by +$1.19\%$ in AP) and on act segmentation on the TRIPOD dataset (by +$5.51\%$ in TA). 
Our contributions are:

\begin{enumerate}[itemsep=0pt,parsep=0pt]
    \item Alignment positional encoding module and a fusion bottleneck layer that performs multimodal fusion with aligned multi-modality inputs.
    \item A cross-domain knowledge transfer module that synchronizes features across-domain, and enables knowledge distillation without requiring extra information.
    \item SoTA performance on scene and act segmentation tasks, with detailed ablations, which can be used as reference for future work.
\end{enumerate}






\section{Related Work}

\textbf{Scene Segmentation in Cinematic Content}:
Recent works on scene segmentation have explored self-supervised learning (SSL)~\cite{chen2021shot,wu2022scene,mun2022boundary}. Self-supervised pretext tasks have included maximizing the similarity between nearby shots compared to randomly selected ones~\cite{chen2021shot}, maximizing the similarity between pairs of images selected according to scene consistency~\cite{wu2022scene}, and maximizing the similarity between pairs of images selected according to pseudo scene boundaries~\cite{mun2022boundary}.
While several previous works have used multimodal inputs for this task~\cite{rao2020local,wu2022scene,chen2021shot}, they have either utilized late fusion of features with predefined weights for each modality~\cite{rao2020local} or have utilized early integration of features derived via SSL~\cite{chen2021shot,wu2022scene}. 
In this paper, we explore how to better align and integrate features from different modalities for scene segmentation.

\textbf{Act Segmentation in Cinematic Content}: Movie screenplays follow a general narrative structure on how a plot unfolds across the story. Several theories have been proposed in this domain dating as far back as Aristotle, who defined a 3 act structure with a beginning (protasis), middle (epitasis), and end (catastrophe)~\cite{pavis1998dictionary}.
Modern screenplays are usually written with a 6 act structure~\cite{hauge2017storytelling}, named as ``the setup", ``the new situation", ``progress", ``complications and higher stakes", ``the final push", and ``the aftermath", separated by five turning points (TPs). 
Prior approaches in narrative segmentation on movies have adopted the aforementioned 6 act structure and posed the problem as identifying the 5 TPs that separate these 6 acts.
~\cite{papalampidi2021movie} is to our knowledge the only prior work that utilizes visual domain in act segmentation by using a pre-trained teacher model trained on textual input to train a student Graph Convolutional network with audio-visual-text representations as input. 
In contrast, our work uses a new multimodal fusion and distillation applied on the modalities which are available with the movie.

\textbf{Multimodal Aggregation}: 
Previous SoTAs on multimodal fusion with transformers perform early fusion of the features as inputs to the transformer~\cite{jaegle2021perceiver}, merge attention between them requiring more memory~\cite{dou2022empirical,kim2021vilt}, use cross attention between two modalities~\cite{wang2020deep,dou2022empirical}, or add cross-modal interactions more efficiently via bottleneck tokens or exchanging some of their latents~\cite{nagrani2021attention,hendricks2021decoupling}. 
~\cite{nagrani2021attention} provides information flow between modalities efficiently by utilizing bottleneck tokens to tame the quadratic complexity of pairwise attention. 
This global exchange between modalities may not be enough for long videos, which require an adaptive fusion in different temporal locations. 
Our model considers~\cite{nagrani2021attention} as baseline and extends it to incorporate local information during information exchange between modalities. 

\textbf{Positional Encoding}:
Previous studies on improving the positional encoding in long sequence modeling have mostly focused on adding relative distance positional encoding ~\cite{shaw2018self,luo2021stable,su2021roformer}. 
However, they do not offer solutions on better maintaining the relative position of latent tokens with respect to their starting point in time in long sequences with variable lengths, which is important for long movie narrative understanding~\cite{hauge2017storytelling,brody2018save}. 
We propose Alignment  Positional Encoding to bridge this gap.

\textbf{Cross-Modality Synchronization \& Distillation}:
To transfer the labels from synopsis to movie shots, we use cross-modality distillation. Previous cross-modality distillation studies for label transfer across modalities are focused on parallel data with the same granularity~\cite{gupta2016cross,aytar2016soundnet,asano2020labelling}, or where the alignment is known~\cite{papalampidi2021film}. 
Alignment of features at different granularities in the same modality, such as screenplay scenes to synopsis sentences~\cite{mirza2021alignarr,zhao-etal-2022-learning} and across modalities such as aligning synopsis sentences to visual shots~\cite{tapaswi2015aligning, xiong2019graph}, book chapters to visual scenes~\cite{tapaswi2015book2movie} have been previously explored. 
While majority of these works rely on unsupervised optimization techniques~\cite{mirza2021alignarr,zhao-etal-2022-learning,tapaswi2015aligning,tapaswi2015book2movie}, there are studies that use supervised labels to improve the representations used for optimization~\cite{xiong2019graph}.
We present an alignment approach with self-supervised loss for synchronizing data in different modalities of cinematic content to enable distillation.

\section{Methodology}
\label{sec:method}
\begin{figure*}[t]
    \centering
    \includegraphics[width=0.93\textwidth]{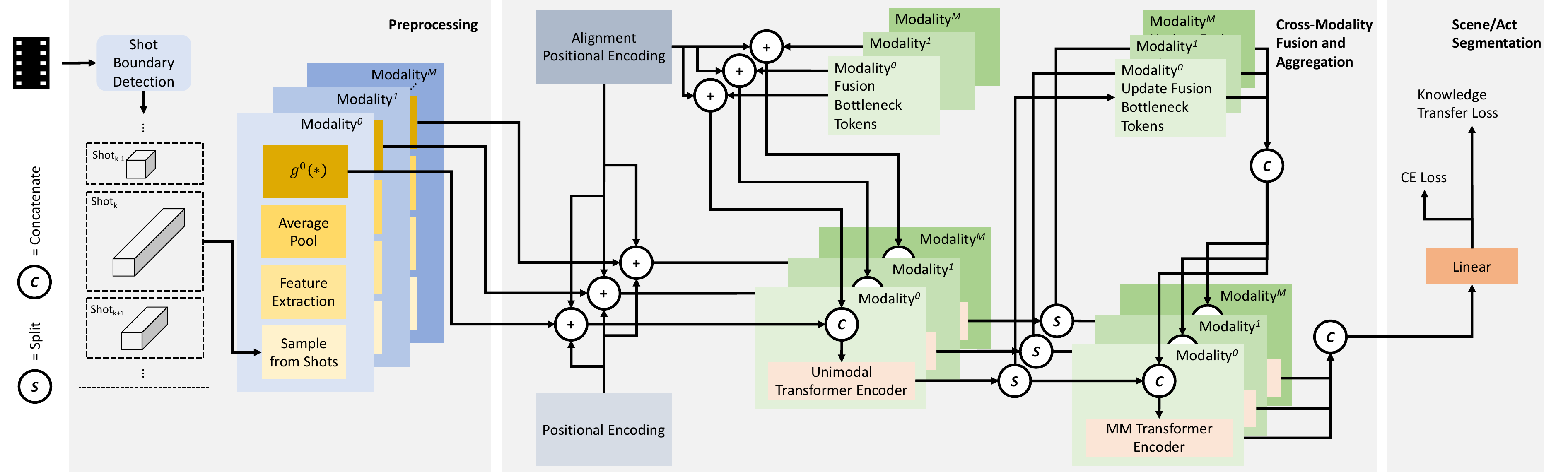}
    \vspace{-5pt}
    \caption{The pipeline of the proposed method includes 1) Preprocessing: splitting the video into shots, extraction of features from each shot, pooling and normalization 2) Cross modality fusion and alignment: with the help of alignment positional encoding and bottleneck fusion tokens, 3) Scene/Act Segmentation comprising the segmentation heads. For Scene Segmentation, CE loss is used and for Act Segmentation knowledge transfer loss is used (refer to Sec.~\ref{sec:distillation} for details)}.
    \label{fig:pipeline}
\end{figure*}

MEGA processes long videos and performs video segmentation in three major steps (Fig.~\ref{fig:pipeline}).
First, a video $V$ is chunked into shots, and multiple features such as scene related features and sound event features are extracted at the shot-level (Sec.~\ref{sec:preprocessing}). The system is built on shot-level representation for two reasons: (1) scene and act boundaries always align with shot boundaries, and (2) the content within a shot remains similar, which allows efficient yet meaningful sampling without losing important information. 
Second, the embeddings from different input samples and each modality are coarsely aligned with a proposed alignment embedding (Sec.~\ref{sec:fuse_aggregation}), and the alignment positional tokens are used to refine the commonly used bottleneck fusion tokens for cross-modal feature fusion (Sec.~\ref{sec:fuse_aggregation}). 
Third, a linear layer is applied on top of the fused representations to generate scene and act boundaries (Sec.~\ref{sec:segmentation}). Finally, 
to address the challenge of cross-domain knowledge transfer where labels from one domain may not directly align with another domain (e.g. act labels on synopsis sentences do not have movie-level timing information), we propose a cross-modal synchronization module that is simple yet effective (Sec.~\ref{sec:distillation}).

\subsection{Preprocessing}
\label{sec:preprocessing}
We chose to utilize Transnet-v2~\cite{souvcek2020transnet} for shot segmentation due to its superior performance and efficiency. 
We list our selection of pre-trained models and associated parameters in Tab.~\ref{tab:feature_extraction}.
As each pretrained feature extraction model has different requirements for input resolution and sampling rate, we first sample the input at various sample rates (as shown in Table~\ref{tab:feature_extraction}). 
It is worth noting that the $\text{CLIP}_{\text{movie}}$ model is the  CLIP~\cite{radford2021learning} model with ViT-B/32 backbone fine-tuned on paired IMDB-image/text dataset. 
IMDB-image dataset comprises 1.6M images from 31.3K unique movies/TV series with 761 unique textual labels. 
The features attributed to each shot are the ones with overlap with the shot time stamp (More details are in Appendix).
After feature extraction, we aggregate the features for each shot and normalize the feature dimension with linear projection as:
\begin{equation}
\label{eq:feature_extract}
  E^m_i=g^m\left(\frac{1}{T^m_i} \sum_{j=1}^{T^m_i} S^m_{ij}\right)
\end{equation}
where $E^m_i \in \mathbb{R}^{C}$ denotes the embedding from the $i$-th shot of $m$-th modality, 
$S^m_{ij} \in \mathbb{R}^{D^m}$ is the $j$-th sampled feature of $m$-th modality from $i$-th shot, and $g^m$  is a linear projection layer that projects feature dimension for $m$-th modality to a common dimension $C$ across all modalities. 

While it is possible to create an end-to-end system starting from raw shot inputs and training the model from scratch, pre-extracting the features from pretrained models is generally more scalable and efficient in actual industrial scenarios, hence we rely on the latter.

\begin{table} 
    \centering
    \scalebox{0.90}{
    \footnotesize
    \begin{tabular}{lllc}
    \hline
    \textbf{Feature extractor} & \textbf{Input Freq.} & \textbf{Input Res.} & \textbf{Feature dim.}   \\
        \toprule
    	\rowcolor{Gray}
            \multicolumn{4}{l}{\textbf{Visual input}} \\
            BASSL~\cite{mun2022boundary} (bassl)            & Varying  & $224^2$    & $2048$ \\
            ResNet$_\text{Place}$~\cite{zhou2017places} (place)     & $1Hz$   &$224^2$ &      $2048$ \\
            ResNeXt101~\cite{xie2017aggregated} (appr)    &$1Hz$   & $224^2$     & $2048$ \\
            I3D~\cite{carreira2017quo} (action)              & $16Hz$   & $16 \times 224^2$    & $2048$ \\
            $\text{CLIP}_{\text{movie}}$ (clip)            & $1Hz$   & $224^2$    & $768$ \\
    	\rowcolor{Gray}
            \multicolumn{4}{l}{\textbf{Acoustic input}} \\
            PANNs~\cite{kong2020panns} (audio) & $1Hz$   & $10 \times 32K$    & $2048$ \\
    	\rowcolor{Gray}
            \multicolumn{4}{l}{\textbf{Linguistic input}} \\
            all-MiniLM-L6-v2~\cite{all-MiniLM-L6-v2-HuggingFace} (text)  & Varying   & -    & $384$ \\
        \bottomrule
    \end{tabular}
}
    \caption{Sampling strategy and feature extraction backbones used for different modalities.}  
    \label{tab:feature_extraction}
    \vspace{-5mm}
\end{table}

\subsection{Cross-modality Alignment and Fusion}
\label{sec:fuse_aggregation}
For long video segmentation into scenes and acts it is important to model short and long term context and perform effective multimodal fusion. However, the commonly used learnable positional embedding only provides fine-grained granularity and is not suitable for high-level semantic alignment across modalities. Furthermore, for tasks such as act segmentation we expect consistent patterns at normalized position of temporal inputs as the theory suggests approximate locations for each turning point (i.e., 10\%, 25\%, 50\%, 75\%, 95\%~\cite{hauge2017storytelling,brody2018save,papalampidi2019movie}). Hence, in  addition to using the traditional positional encoding, we introduce Alignment Positional Encoding layer $\in \mathbb{R}^{L_n \times C}$, which is a learnable embedding layer, for which the index at $i$-th temporal unit (e.g., shot) is derived by:
\begin{equation}
\label{eq:feature_extract2}
  i_{align}=\mathrm{floor}\left({\frac{L_n}{L}i}\right)
\end{equation}
where  $L$ is the temporal dimension (e.g., number of shots) and $L_n$ is the length of alignment positional encoding which is a hyper-parameter ($L_n < L$).
We add the alignment positional encoding to the features in conjunction with the conventional positional encoding (see Fig~\ref{fig:pipeline}).
This module is shared across different modalities. This module provides extra information to the network that can be helpful in learning from long training samples with varying lengths, and in coarsely aligning inputs from different modalities before information fusion.

\begin{figure}
    \centering
    \includegraphics[scale=0.16]{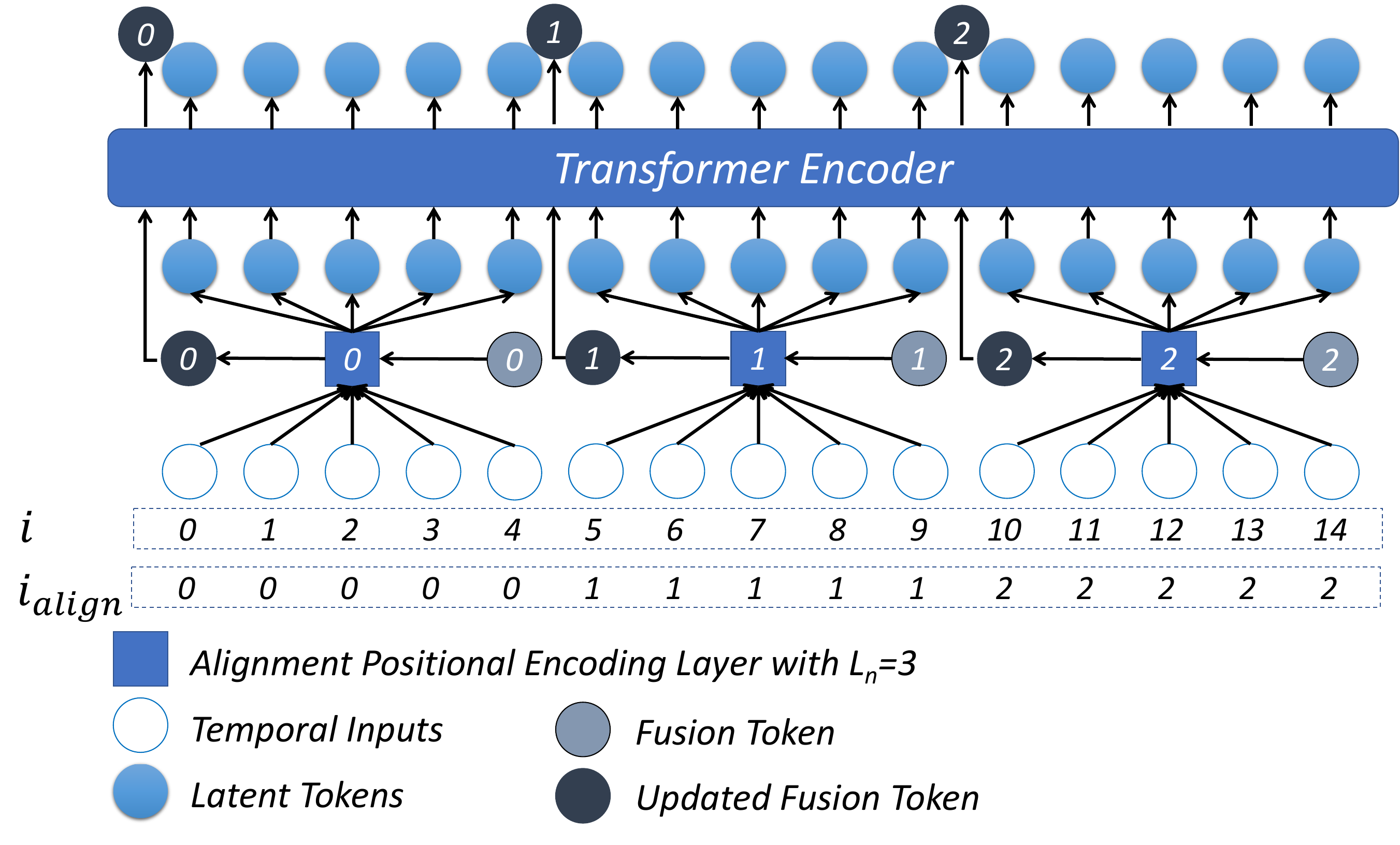}
    \caption{Illustration of Normalized Positional Encoding integration to the temporal tokens in one modality. This figure shows the integration of normalized Positional Encoding with 1) temporal shots for one modality and 2) bottleneck fusion tokens, where $L=15$ and $L_n =3$. For (1) $i_{align}$ is obtained per shot index, $i$, and then each shot is integrated with normalized PE. For (2) each randomly initialized bottleneck token is integrated with normalized PE for its corresponding index.}
    \label{fig:norm_PE}
\end{figure}

Inspired by previous works~\cite{wang2020deep,nagrani2021attention,dou2022empirical}, we choose cross-modal feature fusion, which has been shown to be more effective than early or late fusion. 
To make our approach scale to multiple modalities, we propose an efficient temporal bottleneck fusion based on~\cite{nagrani2021attention}, and follow their mid-fusion strategy, which comprises of an unimodal transformer encoder followed by a multimodal transformer encoder (See Fig.~\ref{fig:pipeline}).
While~\cite{nagrani2021attention} proposes to use bottleneck tokens for fusing information across different modalities, these tokens learn to integrate the information across modalities in a global manner. 
We propose to use $L_n$ fusion tokens, and then integrate them with the same Normalized Positional Encoding to align them with features on the coarse temporal scale (Fig.~\ref{fig:norm_PE}).

The transformer layer per modality then takes in an extra set of aligned fusion tokens concatenated with its input (Fig~\ref{fig:pipeline}), making it much more efficient compared to other methods such as merged attention or pairwise cross-attention with respect to the number of modalities~\cite{dou2022empirical,kim2021vilt}.
Finally, the latent tokens per modality from the last fusion layer (i.e., $Z^m$ for m-th modality) are concatenated as the fused representation:
\begin{equation}
\label{eq:fused_latent}
    Z^{\text{fused}} = \text{concat}_C\left(Z\right)=\text{concat}_C\left(\left[Z^1,...,Z^M\right]\right)
\end{equation}
where $\text{concat}_C(*)$ stands for channel-wise concatenation and $M$ is the number of modalities.

\subsection{Scene/Act Segmentation}
\label{sec:segmentation}
MEGA adopts a similar approach to previous works~\cite{wu2022scene,mun2022boundary,chen2021shot}, where scene segmentation is framed as a binary classification task using a key-shot representation from a window of $2\times k+1$ shots ($k$-shots before and after):
\begin{equation}
    y_i = \varphi\left(Z^{\text{fused}}_i;\theta_s\right)
    \label{eq:sseg_head}
\end{equation}
where $y_i$ is the logit prediction for the $i$-th key-shot. $\varphi$ denotes the linear layer with learnable parameters $\theta_s$ and a cross entropy loss is utilized\cite{wu2022scene,mun2022boundary,chen2021shot}.

The act segmentation task is formulated as $N_{tp}$  linear prediction heads ($N_{tp}$ is the number of turning points, and $\theta_{a_n}$ denotes the head parameters for $n$-th turning point~\cite{papalampidi2021movie}), for each individual shot from a temporal model that takes all the shots of the movie as input. To make a prediction at the $i-$th shot for the $n$-th act boundary ($n \in \left\{1,\dots,N_{tp}\right\}$), we use:
\begin{equation}
    y_{in} = \varphi\left(Z^{\text{fused}}_i;\theta_{a_n}\right)
    \label{eq:aseg_head}
\end{equation}
where $y_{in}$ is the logit prediction for $i$-th shot and $n$-th turning point. 

\subsection{Cross-domain Knowledge Transfer}
\label{sec:distillation}
It is quite common during machine learning that certain modalities may lack annotations that are directly available in other modalities with different information granularity. 
To address this, we propose a cross-modality synchronization scheme that enables cross-modal distillation.
We utilize this module for act segmentation where we aim to transfer act labels from synopsis sentences to movie-level timestamps.
Importantly, our approach does not require additional information, such as screenplay~\cite{papalampidi2021movie}, to bridge the gap.

Our knowledge distillation utilizes (1) an individual network to learn the synopsis-based act segmentation in a supervised manner with cross-entropy loss similar to~\cite{papalampidi2021film}; and (2) a novel synchronization approach between synopsis and movie.
For (1) we use the same architecture mentioned in Secs.~\ref{sec:preprocessing},~\ref{sec:fuse_aggregation}, setting
$C_\text{fused}$ equal to the multimodal shot model setting. 
Additionally, similar to shot level linear prediction head (See Eq.~\ref{eq:aseg_head}), we use a sentence level linear prediction head, resulting in $q_{in}$ logits for the $i$-th sentence of $n$-th TP.
A supervised Cross Entropy loss is used to learn the synopsis labels from predictions for each turning point ($\mathscr{L}_{ce}$).
For (2) we seek a synchronization matrix $W$ $\in \mathbb{R}^{L_{sh} \times {L_{syn}}}$ between $L_{sh}$ shots and $L_{syn}$ synopsis sentences for a sample, where $w_{ij}=1$ if  the $i$-th shot matches with the $j$-th synopsis sentence and $w_{ij}=0$, otherwise. 
Assuming $F(.;\theta)$ represents a parametric reward function (with parameters $\theta$), to find $W$, we define an objective as:
\begin{equation}
\label{eq:em}
\begin{split}
    \underset{W,\theta}{\mathrm{max}} \sum_{i,j} & w_{ij}   F(.;\theta) - \lambda \sum_{i, j} | w_{i,j} |
    \\
    & s.t.\ 0\leq w_{ij} \leq 1
    \end{split}
\end{equation}
Expectation-Maximization algorithm is used to solve the objective in Eq.~\ref{eq:em}. We estimate the target variable $W$ via fixed parameters (i.e., $\theta$) in the E-step, and update the parameters while the target variable is known in the M-step. 

{\noindent \textbf{E-step}:} Assuming $F(.;\theta)$ returns the similarity of input shot and synopsis sentence pair, the E-step has a closed form solution. In the E-step, following~\cite{mirza2021alignarr}, we reduce the search space during optimization to only the pairs which are inside a diagonal boundary (see the proof for E-step and visualization of expected synchronization matrix for different examples in Appendix). 

{\noindent \textbf{M-step}:}
Using all samples in a batch with  $L_{\mathrm{SH}}$, $L_{\mathrm{SYN}}$ total number of shots and synopsis sentences, we form their cosine similarity matrix of dimension $L_{\mathrm{SH}} \times L_{\mathrm{SYN}}$. 
For each query (synopsis sentence/movie shot, respectively), the positive keys (movie shot/synopsis sentence, respectively) are derived from the expectation step. 
Negative keys are the keys lying outside the diagonal boundary of the similarity matrix of shot-synopsis pairs for one movie~\cite{mirza2021alignarr}, and all the keys from other movies within the batch. 
Here each query (synopsis sentence/movie shot, respectively) can have more than one positive key (movie shot/synopsis sentence, respectively) attached to it. 
Following~\cite{khosla2020supervised}, we adopt a modified version of the InfoNCE loss and combine it with the symmetric contrastive loss~\cite{radford2021learning} as:
\begin{equation}
	\label{eq:loss}
    \begin{split}
	\mathscr{L}_{c} =& 
	- \sum_{i=1}^{L_{\mathrm{SYN}}}\frac{1}{|\hat{y}_i|}\sum_{k=1}^{L_{\mathrm{SH}}}\hat{y}_{ik}\mathrm{log}\frac{exp(v_i u_k / \tau)}{\sum_{j=1}^{L_\mathrm{SH}} exp(v_i u_j / \tau)} - \\
	& 
	\sum_{i=1}^{L_{\mathrm{SH}}}\frac{1}{|\hat{y}^T_i|}\sum_{k=1}^{L_{\mathrm{SYN}}}\hat{y}^T_{ik}\mathrm{log}\frac{exp(u_i v_k  / \tau)}{\sum_{j=1}^{L_{\mathrm{SYN}}}exp(u_i v_j / \tau)} 
	\end{split}
\end{equation}
where $\tau$ is a learnable temperature parameter, and $\hat{y}_{ij}$ is a binary indicator of positive vs. negative pairs, $u_i$ is the normalized feature for the $i$-th shot and $v_i$ is the normalized feature for the $i$-th synopsis sentence.

{\noindent \textbf{Knowledge distillation}:}
Knowledge distillation is used to transfer the  knowledge available for the training samples on synopsis. 
The predictions from the synopsis model are mapped to shots using a matrix of their similarities as calculated in the maximization step, for each sample.
The similarity scores for each shot are normalized along the synopsis sentences with softmax. 
The logit predictions from synopsis model are transferred to shots by multiplication with the normalized similarity matrix.  
A softmax along the shots is applied to the transferred logits to derive the probability scores for each shot.
Following~\cite{papalampidi2021film}, we use a Kullback–Leibler divergence loss between predicted outputs for each shot and the transferred probabilities ($\mathscr{L}_{kd}$) (More details are provided in Appendix.).

The cross-domain knowledge transfer module can be trained by simply adding the losses together as: 
\begin{equation}
\label{eq:loss_all}
    \mathscr{L}=\alpha_c \mathscr{L}_{c} + \alpha_{ce}\mathscr{L}_{ce} + \alpha_{kd}\mathscr{L}_{kd}
\end{equation}
where $\alpha_c$, $\alpha_{ce}$, and $\alpha_{kd}$ are hyperparameters that control the weights of the three losses.

\section{Experiments}

\subsection{Dataset}
\label{sec:datasets}
We test our model on two commonly used dataset: 
\noindent\textbf{Movienet-318~\cite{huang2020movienet}: } consists of 1100 movies, out of which 318 movies are annotated for the task of scene segmentation. 
The annotated dataset is split into 190, 64, and 64 movies for train, validation and test splits, respectively. 
We report the Average Precision (AP) and F1-score (F1) on the test split following previous  work~\cite{rao2020local,wu2022scene}.

\noindent\textbf{TRIPOD~\cite{papalampidi2021movie}} includes 122 movies, split into 84, and 38 movies for train and test, respectively. 
This dataset includes the annotations of 6 act boundaries (``the setup", ``the new situation", ``progress", ``complications and higher stakes", ``the final push", ``the aftermath") on the movie synopsis sentences for the training set, and on the movie screenplay scenes for the test set. 
The authors also have released soft probability scores (silver annotations) for the training set, using~\cite{papalampidi2019movie}\footnote{\url{https://github.com/ppapalampidi/GraphTP}}. 
To find the timestamps for the screenplay scenes in the movie, following~\cite{papalampidi2021movie} we used Dynamic Time Warping (DTW) to align the timed subtitles from the movie to the monologue lines in the screenplay. 
Following~\cite{papalampidi2021movie}, we use total agreement (TA), partial agreement (PA) and distance (D) as evaluation metrics. 

\subsection{Implementation Details}
\label{sec:implementation_details}
\noindent\textbf{For scene segmentation: }
We train our model with 8 V100 GPUs with total batch size of 1024. 
The Adam~\cite{kingma2014adam} optimizer is used with  learning rate of 1e-4.
We train the model for 20 epochs.
GeLU~\cite{hendrycks2016gaussian} is used as activation function by default, we use  weighted cross entropy to balance the positive and negative samples at batch level.
We choose shot sequence length of 17 ($k=8$) following~\cite{mun2022boundary}. 
We set $L_n=2$ for this model.  

\noindent\textbf{For act segmentation: }
We train our model with 4 V100 GPUs with total batch size of 4. 
The SGD optimizer is used with learning rate of 1e-3. We train the model for 10 epochs.
$\lambda$ in Eq.~\ref{eq:em} is empirically set differently for each synopsis sentence of each sample, by finding 99\% percentile of the similarity scores between the synopsis sentence and all the shots corresponding to that sample.
$\alpha_c$, $\alpha_{ce}$, $\alpha_{kd}$ are set to 1, 1, and 10. We set $L_n=100$ for this the shot model and $L_n=20$ for the synopsis model. 
We use max pooling to aggregate the shot-level predictions to scene level.
We use all shots from a video for act segmentation.

\subsection{Main Results}
\label{sec:exp}

\noindent\textbf{SoTA on Scene Segmentation.}
We first show MEGA outperforms previous SoTA on MovieNet318~\cite{mun2022boundary} for scene segmentation (+$1.19\%$ on AP and +$8.28\%$ on F1). 
With the same input visual features, MEGA outperforms previous SoTA~\cite{mun2022boundary}  (Tab.~\ref{tab:sota_ss} +$0.52\%$ on AP and +$3.69\%$ on F1), which indicates that the proposed approach is effective. 
Thanks to the proposed cross-modality fusion module, the MEGA generalizes and benefits from additional information extracted from visual signals.  
MEGA with 3 visual modalities (clip, place, and bassl) outperforms single modality model by +$0.67\%$ on AP and +$4.59\%$ on F1, which shows the proposed  fusion works as expected. 
It is worth mentioning that the proposed method is scalable and generalizes to various number of modalities at different scales, which makes it flexible for real-world applications.

\begin{table} 
    \centering
    \scalebox{0.97}{
    \footnotesize
    \begin{tabular}{lllcc}
    \hline
    \multirow{2}{*}{\textbf{\textit{Approach}}} & \multirow{2}{*}{\textbf{\textit{Modality}}} & \multirow{2}{*}{\textbf{\textit{Pretrained on}}} & \textbf{\textit{AP $\uparrow
$}} & \textbf{\textit{F1$\uparrow
$}}   \\
          &  &&  \textbf{\textit{[\%]}} & \textbf{\textit{[\%]}}\\
         \hline
            Random~\cite{rao2020local}& - & - & 8.2 & - \\
            \hline
            \multicolumn{5}{l}{\textbf{Visual only input}}
            \\
            GraphCut~\cite{rasheed2005detection} &V & Places~\cite{zhou2017places}& 14.1 & - \\
            SCSA~\cite{chasanis2008scene} &V&Places~\cite{zhou2017places}& 14.7 & - \\
            DP~\cite{han2011video} &V&Places~\cite{zhou2017places}& 15.5 & - \\
            Grouping~\cite{rotman2017optimal} &V&Places~\cite{zhou2017places}& 17.6 & - \\
            StoryGraph~\cite{tapaswi2014storygraphs}&V&Places~\cite{zhou2017places}& 25.1& -\\
            Siamese~\cite{baraldi2015deep} &V&Places~\cite{zhou2017places}& 28.1 & -\\
            LGSS~\cite{rao2020local} &V & Places~\cite{rao2020local}& 39.0 & - \\
            LGSS~\cite{rao2020local} &V & Cast~\cite{huang2018unifying,zhang2015beyond}& 15.9 & - \\
            LGSS~\cite{rao2020local} &V & Action~\cite{gu2018ava}& 32.1 & - \\
            ShotCoL~\cite{chen2021shot}\textdagger & V &Movienet~\cite{huang2020movienet}& 52.89 & 49.17\\
            SCRL~\cite{wu2022scene} & V &Movienet~\cite{huang2020movienet}& 54.82	& 51.43\\
            BaSSL~\cite{mun2022boundary} &V &Movienet~\cite{huang2020movienet} & 57.4 
            & 47.02 
            \\
        	\rowcolor{Gray}
            \textbf{MEGA} & V  &Movienet~\cite{huang2020movienet}& 57.92 &  50.71 \\
        	\rowcolor{Gray}
            \textbf{MEGA}& 
            V&M+P+I& 
            58.59&
            55.30
            \\
            \hline
    \end{tabular}
}
\vspace{-0.1cm}
    \caption{Scene boundary detection: comparison with SoTA. \textdagger means the numbers are copied from~\cite{wu2022scene}. M+P+I denotes pre-trained on Movienet~\cite{huang2020movienet}, Places~\cite{zhou2017places} and IMDB.
    }
    \label{tab:sota_ss}
    \vspace{-4mm}
\end{table}

\noindent\textbf{SoTA on Act Segmentation.}
We then show that MEGA establishes the new SoTA performance on act segmentation on TRIPOD dataset (Tab.~\ref{tab:sotatpd}).  
We first show that MEGA outperforms previous SoTA on TRIPOD~\cite{papalampidi2021movie} dataset With only visual signals as input.
Comparing to other works that take textural inputs~\cite{papalampidi2020screenplay,papalampidi2021movie}, MEGA is able to achieve better performance. Furthermore, in real-world applications, the visual input (the video) is often easier to obtain than textual inputs such as screenplay~\cite{papalampidi2021movie}.
We further show that MEGA* (Tab.~\ref{tab:sotatpd}), which swaps our synchronization module to use similar synchronization as SoTA~\cite{papalampidi2021movie}, outperforms GRAPHTP, which demonstrates the effectiveness of proposed approach including the alignment and fusion modules. 
It is worth mentioning that, with multiple features extracted from \textbf{visual} media modality alone, MEGA outperforms previous SOTA which makes the proposed model applicable for real-world scenarios, as it is usually harder to get additional media modalities (e.g., screenplay) which are used in other works.

By further aggregating the results from text input, MEGA establishes the new state-of-the-art on TRIPOD dataset (Tab.~\ref{tab:sotatpd}). 
MEGA almost doubles the performance of previous work~\cite{papalampidi2020screenplay,papalampidi2021movie} 
with +$5.51\%$ TA, +$9.15\%$ PA, -$\%0.81$ D.
This shows the proposed multimodal fusion scales and generalizes well to multiple modalities. 
The cross-modality distillation also works robustly for various settings.
We noticed that adding the acoustic features is not always helpful to the performance (Tab.~\ref{tab:sotatpd}), probably because acoustic information provides redundant or not useful information for the task of act boundary segmentation. 

\begin{table} 
\centering
\scalebox{0.80}{
    \centering
    \small
    \begin{tabular}{lllccc}
    \hline
    \multirow{2}{*}{\textbf{\textit{Approach}}} & \multirow{2}{*}{\textbf{\textit{Modality}}}& \textbf{\textit{Modality}}  & \textbf{\textit{TA$\uparrow
$}}  & \textbf{\textit{PA$\uparrow
$}} & \textbf{\textit{D$\downarrow
$}}  \\
          &  & \textbf{\textit{for synch.}} & \textbf{\textit{[\%]}} & \textbf{\textit{[\%]}} & \textbf{\textit{[\%]}} \\
        \hline 
    Random (Evenly dist.)~\cite{papalampidi2021movie}& - & T*&4.82 &6.95& 12.35 \\
    Theory~\cite{hauge2017storytelling,papalampidi2019movie}& - & T*& 4.41 &6.32  & 11.03\\
    Distribution position~\cite{papalampidi2021movie}& - & T*&5.59&  7.37 & 10.74\\
    \hline
    \multicolumn{4}{l}{\textbf{Single modality input}} \\
 
    TEXTRANK~\cite{mihalcea2004textrank}& T* & T*& 6.18 & 10.00 & 17.77 \\
    SCENESUM~\cite{gorinski2015movie}& T* & T* & 4.41 & 7.89&  16.86\\
    TAM~\cite{papalampidi2020screenplay}& T* & T*& 7.94 & 9.47 & 9.42\\
    GRAPHTP~\cite{papalampidi2021movie} & T* & T*& 6.76 & 10.00 & 9.62 \\  
	\rowcolor{Gray}
    MEGA* & V & T* & 10.51 & 14.54  & 8.98 \\
	\rowcolor{Gray}
    MEGA & V & V & 13.93  & 20.72 & 9.19  \\
    \hline
    \multicolumn{4}{l}{\textbf{Multi-modality input}} \\
    TEXTRANK~\cite{mihalcea2004textrank}& T*+A+V& T* & 6.18 & 10.00 & 18.90\\
    SCENESUM~\cite{gorinski2015movie}& T*+A+V & T*&6.76  & 11.05 &18.93 \\
    TAM~\cite{papalampidi2020screenplay}& T*+A+V & T*& 7.36 & 10.00 & 10.01\\
    GRAPHTP~\cite{papalampidi2021movie} & T*+A+V& T*& 9.12 & 12.63 & 9.77 \\
	\rowcolor{Gray}
    MEGA* & T+V & T* & 11.14 & 15.20 & \textbf{8.96  } \\
	\rowcolor{Gray}
    MEGA & T+V  & T+V &  \textbf{14.63 } &   21.78  &  \textbf{8.96 } \\
	\rowcolor{Gray}
    MEGA* & T+A+V& T*  & 10.00  &  14.08&  8.96 \\
	\rowcolor{Gray}
    MEGA & T+A+V & T+A+V&  14.19 &   \textbf{22.10} &  9.68\\
    \hline
    \end{tabular}
}
    \caption{TP identification: comparison with SoTA. MEGA* denotes the MEGA using the same synchronization as~\cite{papalampidi2019movie} for fair comparison. T*,V,T,A denote Textual-screenplay, Visual, Textual-subtitle and Acoustic features, respectively.}
     \label{tab:sotatpd}
     \vspace{-5mm}
\end{table}

\subsection{Ablations}
We perform ablations to examine the effectiveness of major building blocks of MEGA. 
We use features extracted from visual media modality (for scene segmentation: clip, place, bassl and for act segmentation: clip, appr, action, place), with the proposed normalized positional encoding, multi-modality bottleneck fusion, and cross-modality synchronization by default unless specified. 
The training and evaluation follows same protocols as mentioned in Sec.~\ref{sec:exp}.

\begin{table}[t]
\footnotesize
\centering
    \subfloat[Effectiveness of Alignment Positional Encoding.]
    {
    \scalebox{1}{
        \begin{tabularx}{0.45\textwidth}{lcccc}
        \toprule
        & \multicolumn{1}{c}{\textit{\textbf{Scene Seg.}}} & \multicolumn{3}{c}{\textit{\textbf{Act Segmentation}}}   \\ 
		\cmidrule(lr){2-2} \cmidrule(lr){3-5} 
        {\textit{case}}  &  {\textit{AP$\uparrow
$}} & {\textit{TA$\uparrow
$}}&  {\textit{PA}$\uparrow$} & {\textit{D$\downarrow
$}}\\
            \midrule
              w/o align. PE          & 57.77 & 5.29 & 7.37 & 31.04\\
              w. align. PE             & \textbf{58.59} & \textbf{13.93} & \textbf{20.72} & \textbf{9.19}\\
              
        \bottomrule	
        \end{tabularx}
        }
        \label{tab:norm_PE}
    }\hfill
    \subfloat[Effectiveness of Normalized Positional Encoding in bottleneck tokens.]
    {
    \scalebox{1}{
        \begin{tabularx}{0.45\textwidth}{llcccc}
        \toprule
        & & \multicolumn{1}{c}{\textit{\textbf{Scene Seg.}}} & \multicolumn{3}{c}{\textit{\textbf{Act Segmentation}}}   \\ 
		\cmidrule(lr){3-3} \cmidrule(lr){4-6} 
        {\textit{case}} & {\textit{modality}}  &  {\textit{AP$\uparrow
$}} & {\textit{TA$\uparrow
$}}&  {\textit{PA$\uparrow
$}} & {\textit{D$\downarrow
$}}\\
            \midrule
            w/o align.PE   & V        & 58.31 & 13.60 & 20.53 & 9.47\\
            w. align.PE    & V      & \textbf{58.59} & 13.93 & 20.72 & 9.19\\
            w/o align.PE   & V + T        & - & 13.01 & 20.13 & 9.56\\
            w. align.PE    & V + T    & - & \textbf{14.63} & \textbf{21.78} & \textbf{8.96}\\
        \bottomrule	
        \end{tabularx}
        }
        \label{tab:norm_PEFT}
    }\hfill
    \subfloat[Multi-modal fusion strategies.]
    {
    \scalebox{1}{
        \begin{tabularx}{0.44\textwidth}{lcccc}
        \toprule
        & \multicolumn{1}{c}{\textit{\textbf{Scene Seg.}}} & \multicolumn{3}{c}{\textit{\textbf{Act Segmentation}}}   \\ 
		\cmidrule(lr){2-2} \cmidrule(lr){3-5} 
        {\textit{MM. integ. type}}  &  {\textit{AP$\uparrow
$}} & {\textit{TA$\uparrow
$}}&  {\textit{PA$\uparrow
$}} & {\textit{D$\downarrow
$}}\\
            \midrule
              LateFusion& 58.24 & 12.57 & 19.21 & 10.00 \\
              Bottleneck             & \textbf{58.59}& \textbf{13.93} & \textbf{20.72} & \textbf{9.19} \\
        \bottomrule	
        \end{tabularx}
        }
        \label{tab:MM_fusion}
    }\hfill
    \subfloat[Impact from input modalities on scene seg.]
    {
    \scalebox{0.9}{
        \begin{tabularx}{0.16\textwidth}{lc}
        \toprule
        {\textit{change}}  &  {\textit{AP$\uparrow
$}} \\
            \midrule
              -clip                   & 58.09 \\
              -place            & 57.51 \\
              -bassl & 51.88 \\
              
              -clip-place            & 57.92 \\
            - & \textbf{58.59}    \\
            & \\
        \bottomrule	
        \end{tabularx}
        }
        \label{tab:modality_scene}
    }\hfill
    \subfloat[Impact from input modalities on act segmentation.]
    {
    \scalebox{0.9}{
        \begin{tabularx}{0.26\textwidth}{lccc}
        \toprule
        {\textit{change}}  &  {\textit{TA$\uparrow
$}} & {\textit{PA$\uparrow
$}} & {\textit{D$\downarrow
$}} \\
            \midrule
              -clip                   & 6.09 &  10.66 & 21.81\\
              -place            & 13.57 & 19.87 & 9.22 \\
              -action            & 13.31 & 20.20 & 10.38 \\
              -appr            & 13.42 & 20.59 & \textbf{8.85} \\
              - & 13.93 & 20.72 & 9.19\\
              +subtitle & \textbf{14.63} & \textbf{21.78} & 8.96 \\
        \bottomrule	
        \end{tabularx}
        }
        \label{tab:modality_act}
    }\hfill
    \subfloat[Impact of Synchronization with multimodal video features on act segmentation.]
    {
    \scalebox{1}{
        \begin{tabularx}{0.43\textwidth}{llccc}
        \toprule
        {\textit{synopsis synch. by}}  &{\textit{M for synch.}}  & {\textit{TA$\uparrow
$}}&  {\textit{PA$\uparrow
$}} & {\textit{D$\downarrow
$}}\\
            \midrule
               ~\cite{papalampidi2019movie}  & T*  &  10.51 & 14.54& 8.98 \\
               MEGA & V         & 13.93 & 20.72 & 9.19 \\
               MEGA & V + T       & \textbf{14.63} & \textbf{21.78} & \textbf{8.96} \\
        \bottomrule	
        \end{tabularx}
        }
        \label{tab:align}
    }\hfill 
    \subfloat[Impact from feature set and model size on scene seg. SPS denotes $\#$ of samples per second.]
    {
    \scalebox{0.9}{
        \begin{tabularx}{0.52\textwidth}{llccc}
        \toprule
        {\textit{Approach}}  & {\textit{Feature Set Pretrained on}}  &  {\textit{AP$\uparrow
$}}  &  {\textit{Params$\downarrow
$}}   &  {\textit{SPS$\uparrow
$}}\\
            \midrule
            BaSSL~\cite{mun2022boundary}         &    Movienet      & 57.4 & \textbf{15.77M} & \textbf{6244.99} \\
              LGSS~\cite{rao2020local}         &   M+P+I        & 52.93 & 66.16M & 206.36 \\
              MEGA & M+P+I & \textbf{58.59} & 67.57M & 1736.13 \\
        \bottomrule	
        \end{tabularx}
        }
        \label{tab:feat_size_scene}
    }\hfill
    \subfloat[Impact from feature set and model size on act seg. Set1 includes Visual (appr), Audio (YAMNet), Textual (script-USE). Set2 has Visual (appr,clip,action,place), Audio (audio), Textual (text from subtitle).]
    {
    \scalebox{0.9}{
        \begin{tabularx}{0.52\textwidth}{llccccc}
        \toprule
        {\textit{Approach}} & {\textit{Feature Set}}   &  {\textit{TA$\uparrow
$}} & {\textit{PA$\uparrow
$}} & {\textit{D$\downarrow
$}} & {\textit{Params$\downarrow
$}} & {\textit{SPS$\uparrow
$}} \\
            \midrule
            GRAPHTP~\cite{papalampidi2021movie}             &  Set1~\cite{papalampidi2021movie}    & 9.12 &  12.63 & 9.77 & \textbf{0.745M} & \textbf{25.40} \\
            GRAPHTP~\cite{papalampidi2021movie}             &  Set2    & 4.72 &  7.37 & 9.69 & 6.78M & 14.36 \\
            MEGA             &  Set2    & \textbf{14.19} &  \textbf{22.10} & \textbf{9.68} & 6.78M & 18.24 \\
        \bottomrule	
        \end{tabularx}
        }
        \label{tab:feat_size_act}
    }
	\label{tab:ablation}
	\vspace{-0.3cm}
    \caption{Ablation studies on MEGA components.}
	\vspace{-6mm}
\end{table}

\noindent\textbf{Alignment Positional Encoding.}
We report the ablations on Alignment PE in Tab.~\ref{tab:norm_PE}. 
We notice that the proposed alignment PE consistently improves the performance on both scene and act segmentation tasks, while act segmentation benefits more significantly from it. 
This is because of two reasons: 1) inputs to the act segmentation model have variable lengths (all the shots from a video) as opposed to scene segmentation which has fixed length inputs, hence the alignment PE adds more information to the act segmentation model; and 2) Alignment PE is shared across all modalities and fusion layer and its absence causes more harm to sequences with longer lengths, as  act segmentation takes longer shot sequences (entire video) compared to scene segmentation (17 shots). Overall, the drop in performance indicates that the proposed Alignment P.E. is an essential component for video segmentation tasks.

We further remove the proposed Alignment PE from bottleneck fusion tokens and show results in Tab.~\ref{tab:norm_PEFT}. We observe a performance drop when removing the Alignment PE. It is worth mentioning that the performance drop is more noticeable when multiple modalities are involved, (e.g. V + T model), which suggests that the information from subtitles requires precise temporal alignment in order to be effective during multimodal fusion.

\noindent\textbf{Multimodal Fusion Strategies.}
Tab.~\ref{tab:MM_fusion} compares our proposed fusion tokens with the commonly used late fusion strategy. 
The results show that the temporal bottleneck fusion clearly outperforms late fusion, demonstrating the effectiveness of aligned bottleneck tokens in improving the performance across both tasks.

\noindent\textbf{Different Input Modalities.}
We study the impact of different modalities by removing them from the input and present results on scene segmentation and act segmentation.
In \textbf{scene segmentation (Tab.~\ref{tab:modality_scene})}, we find that the bassl feature followed by place are the most important. This is because the BaSSL\cite{mun2022boundary} model is pretrained for scene segmentation and place consistency is critical for scene segmentation.
In \textbf{act segmentation (Tab.~\ref{tab:modality_act})},  we find that the CLIP feature pre-trained on IMDB, followed by subtitle are the most important features.
This is because the clip model is pre-trained on abstract concepts (e.g. genre, micro genre, character type and coarse key places), thus the CLIP feature contains richer semantics, and subtitle provides complementary rich semantic information. These high level semantics are considered useful for act segmentation. 
Overall, when all the features are included, the model is able to leverage the unique information provided by each feature and yields the best performance. 

\noindent\textbf{Cross-modality Synchronization.}
Tab.~\ref{tab:align} studies the effectiveness of the proposed cross modality synchronization on act segmentation.
To establish a baseline, we use the probability scores provided by~\cite{papalampidi2021movie} and derived by aligning synopsis sentences to scenes using screenplay~\cite{papalampidi2019movie} (T* in Tab.~\ref{tab:align} denotes screenplay). 
For a fair comparison, we repeat the scores provided for each scene on all of its shots and then re-normalize\footnote{This strategy maintains the rank of probability scores for different turning points across different segments of the movie, and is consistent with the max-pooling of prediction scores on shot level to derive the scene level predictions during evaluation (see Sec. \ref{sec:implementation_details}).}. 
MEGA with visual only input outperforms~\cite{papalampidi2019movie} across two metrics (PA and TA) and when we add the subtitle features (T in Tab.~\ref{tab:align} denotes subtitle), MEGA outperforms~\cite{papalampidi2019movie} across all the metrics. 
The results demonstrate that the proposed cross-domain synchronization works effectively and generalize well to various modalities. 
It is worth mentioning that the proposed method generates act segmentation without requiring the screenplay, which makes it practical for various industrial applications. 

\noindent\textbf{Impact of Feature Set.}
In Tab.~\ref{tab:feat_size_scene}, we examine whether feature set plays an outsized role in MEGA's improved performance over other methods. On the same feature set of M + P + I, we outperform LGSS~\cite{rao2020local} while only introducing a small number of additional parameters. BaSSL~\cite{mun2022boundary} achieves slightly lower performance using only Movienet features, indicating that the use of additional features is not the primary reason for our improved performance.


\noindent\textbf{Impact of Model Size.}
We finally look into the impact of model size on the performance. 
To establish the fair comparison, we first expand GRAPHTP~\cite{papalampidi2021movie} which shares the same input as MEGA to roughly the same number of parameters as MEGA.  
Our results on act segmentation (Tab.~\ref{tab:feat_size_act}) show that  MEGA outperforms the previous SoTA, GRAPHTP~\cite{papalampidi2021movie}, which indicates the MEGA has efficient and effective design.

\vspace{-.5mm}
\subsection{Fusion with Audio Modality}
We also experimented with the effect of adding audio. Movienet~\cite{huang2020movienet} has released \emph{Short Time Fourier Transform} (STFT) features extracted from audio files. 
We use the same audio backbone as~\cite{rao2020local}. However, by adding audio features in MEGA, we see a drop in the performance (see Tab.~\ref{tab:limitation}). 
Although~\cite{rao2020local,chen2021shot} have shown improvements across multiple models by adding the audio modality across Movienet-150 dataset~\cite{rao2020local} (where the split is not publicly available) or a private dataset: AdCuepoints~\cite{chen2021shot}, Wu \etal~\cite{wu2022scene} have observed a similar trend as our experiments, where adding the released audio features in Movienet~\cite{huang2020movienet} to SCRL~\cite{wu2022scene} and ShotCoL~\cite{chen2021shot} drops the performance (see Tab.~\ref{tab:limitation}). 
Possible reasons can be 1) the audio features published by Movienet 
via STFT\footnote{\url{https://github.com/movienet/movienet-tools}} are an incomplete view of the shot from audio modality either in representation or in terms of the audio chunk from each shot they used, and the raw audio files are not available. 
2) our multimodal fusion strategy cannot exploit the possible complementary information or filter the harmful or confusing signals from the audio modality.

\begin{table} 
    \centering
    \small
    \begin{tabularx}{0.36\textwidth}{llc}
        \hline

    \multirow{1}{*}{\textit{\textbf{Approach}}} & \multirow{1}{*}{\textit{\textbf{Modality}}}  & \textit{\textbf{AP$\uparrow$}} [\%] \\
         \cmidrule(lr){1-3} 
            LGSS~\cite{rao2020local}&V(place) &39.00\\
            ShotCoL~\cite{chen2021shot}\ddag & V & 46.77\\
            SCRL~\cite{wu2022scene} & V & 54.55	 \\
        	\rowcolor{Gray}
        	\rowcolor{Gray}
            \textbf{MEGA} & V(place,clip,bassl)  & 58.59 \\
            \cmidrule(lr){1-3} 
            LGSS~\cite{rao2020local}&V(place)+A&43.4\\
            ShotCoL~\cite{chen2021shot}\ddag & V+A & 44.32\\
            SCRL~\cite{wu2022scene} &V+A& 50.80  \\
        	\rowcolor{Gray}
            \textbf{MEGA} & V(place,clip,bassl)+A  &  55.36   \\

    \hline
    \end{tabularx}
    \caption{Scene Seg. with audio.  
    \ddag denotes copying from~\cite{wu2022scene}.
    }
    \label{tab:limitation}
    \vspace{-5mm}
\end{table}

\vspace{-1mm}
\section{Discussion and Conclusion}

\noindent\textbf{Limitations.} The explorations in this work are limited to  appearance, location, activity, acoustic and textual features. 
For long movie segmentation, however,  providing the name of actors (tabular data) and having a specific component for actor identification in the movie can help both the synchronization and the act/scene segmentation models. 
We will explore the use of this data.

The results demonstrated that richer semantic representations from the clip features enhanced the performance for long video segmentation. 
To obtain better performing representations for long video understanding one can use large amount of unlabeled data with carefully selected pretext tasks for understanding long context.
We will investigate the use of SSL to train a rich multimodal representation from videos and will examine the learned representations across multiple long video understanding tasks.

\noindent\textbf{Conclusion.} This paper introduces MEGA, a unified solution for long video segmentation.
Our design of normalized positional encoding, and their integration into fusion tokens allows MEGA  to learn consistent patterns from inputs with variable lengths and efficiently and effectively align and fuse them across different modalities. 
Our synchronization schema further allows the use of rich multimodal tokens to be used in transferring the labels from synopsis sentences to movie shots, facilitating the knowledge distillation from synopses to movies. 
MEGA achieves state-of-the-art performance compared with previous works. 



\begin{appendices}

\begin{figure*}[t]
    \centering
    \includegraphics[width=0.8\textwidth]{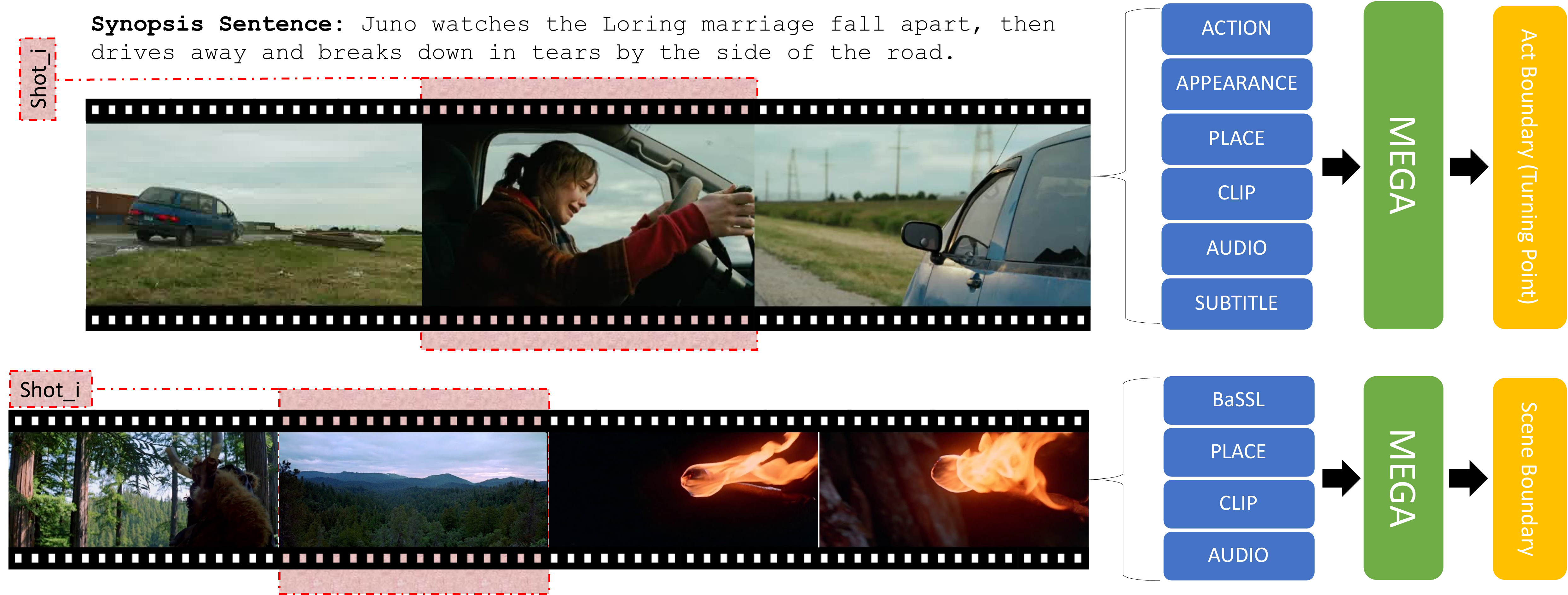}
    \caption{This figure demonstrates the two applications used to test our model for video segmentation: 1) act boundary segmentation (tops) and 2) scene boundary segmentation (bottom). The model utilizes several features extracted from pretrained models to predict a decision for each shot.}
    \label{fig:teaser}
    \vspace{-5mm}
\end{figure*}

\begin{table}[h]
\footnotesize
    \centering
    \subfloat[Architecture]
    {
    \label{tab:arch}
    \scalebox{0.95}{
    \begin{tabularx}{0.51\textwidth}{l|l|ll}
        \cmidrule(lr){2-4}
         &stage & layer & output size\\
         \cmidrule(lr){1-4}
         \multirow{1}{*}{\rotatebox[origin=c]{90}{Unimodal}} &Shot Encoder & Linear: $D^m \times C$  & $L \times C$\\
         &(Sentence& Pos. Enc.: $L \times C$  & $L \times C$\\
         &Encoder)& Align.PE: $L_n \times C$  & $L \times C$\\
         && LayerNorm: $C$ & $L \times C$\\
         && Dropout (p) & $L \times C$\\
         \cmidrule(lr){2-4}
        & Bottleneck & $L_n \times C$  & $L_n \times C$\\
        & & Align.PE: $L_n \times C$  & $L_n \times C$\\
         \cmidrule(lr){2-4}
        & Concat   & & $(L_n + L) \times C$\\
         \cmidrule(lr){2-4}
        & \multirow{7}{*}{\rotatebox[origin=c]{90}{Trans.Enc. $\times N_u$}}&Self Attn: $3 \times C$ & $(L_n + L) \times C$\\
        & & LayerNorm: $C$ & $(L_n + L) \times C$\\
        & & Linear: $C \times C_k$ & $(L_n + L) \times C_k$\\
        & & Activation: GeLU &$(L_n + L) \times C_k$\\
        & & Linear: $C_k \times C$ & $(L_n + L) \times C$\\
        & & LayerNorm: $C$ & $(L_n + L) \times C$\\
        & & Dropout (p)& $(L_n + L) \times C$\\
         \cmidrule(lr){1-4}
        \parbox[t]{2mm}{\multirow{9}{*}{\rotatebox[origin=c]{90}{Fusion}}}&  Concat & & $(M \times L_n + L) \times C$\\
         \cmidrule(lr){2-4}
         
        & \multirow{7}{*}{\rotatebox[origin=c]{90}{Trans.Enc.  $\times N_f$}}&Self Attn: $3 \times C$ & $(M \times L_n + L) \times C$\\
        & & LayerNorm: $C$ & $(M \times L_n + L) \times C$\\
        & & Linear: $C \times C_k$ & $(M \times L_n + L) \times C_k$\\
        & & Activation: GeLU &$(M \times L_n + L) \times C_k$\\
        & & Linear: $C_k \times C$ & $(M \times L_n + L) \times C$\\
        & & LayerNorm: $C$ & $(M \times L_n + L) \times C$\\
        & & Dropout (p)& $(M \times L_n + L) \times C$\\
         \cmidrule(lr){2-4}
        & Concat & & $L \times C_\text{fused}$\\
         \cmidrule(lr){1-4}
        & Output & Linear ($C_\text{fused} \times N_c$) & $L \times N_c$\\
    \end{tabularx}
    }
    }
    \hfill
    \subfloat[Hyperparameters]{
    \label{tab:hyperparams}
        \scalebox{0.95}{
            \begin{tabularx}{0.45\textwidth}{lccc}
            \toprule
            \multirow{2}{*}{Name} & \multirow{2}{*}{Scene Segmentation} & \multicolumn{2}{c}{Act Segmentation} \\
            \cmidrule(lr){3-4}
            & & Shot Model & Syn. Model\\
            \cmidrule(lr){1-4}
            $L$ & 17 & 3000 & 60\\
            $L_n$ & 2 & 100 & 20\\
            $C$ & 768 & 128 & $128 \times M$\\
            $C_k$ & 3072 & 128 & $128 \times M$ \\
            $N_u$ & 2 & 1 & 1\\
            $N_f$ & 1 & 1 & - \\
            $p$ & 0.1 & 0.5 & 0.1\\
            $N_c$ & 2 & 5 & 5\\
            \bottomrule
            \end{tabularx}
        }
    }
    \caption{Architecture details.}
    \label{tab:arch_details}
\end{table}

\section{Architecture}

The details of our model architecture are shown in Tab.~\ref{tab:arch}. The model is composed of unimodal encoders, fusion encoders and output layer. The unimodal encoders are repeated for all $M$ modalities and consist of a Shot Encoder, Bottleneck layer, and Transformer Encoder repeated $N_u$ times. The Fusion starts with concatenating the fusion tokens and further includes the Transformer Encoder repeated $N_f$ times for each modality, and final concatenation of the latent tokens for different modalities ($C_\mathrm{fused} = C \times M$) per shot. The hyperparameter values used to derive the architecture per task (Scene/Act Segmentation) are provided in Tab.~\ref{tab:hyperparams}.

\section{Feature Extraction}
$\text{CLIP}_{\text{movie}}$ model is the the original CLIP~\cite{radford2021learning} model with ViT-B/32 backbone fine-tuned on IMDB-image dataset. 
The IMDB-image dataset includes 1.6M images from 31.3K unique movies/TV series paired with 762 unique textual labels. This model is trained with contrastive loss similar to CLIP~\cite{radford2021learning}. The differences with~\cite{radford2021learning} are: a) the textual labels are from a limited set, b) the positive and negative keys for a query sample are identified by their labels, c) the number of positive keys per image can be more than one in a batch, and d) not all other samples in a batch are considered negative keys for a query sample, only the ones with different sets of labels are considered negative keys.

\section{Details of Feature Extraction Per Shot}
Bassl features are used for scene segmentation and are extracted from 3 key frames per shot, as Movient only releases 3 key frames per shot. Appr, place, clip, action features are extracted every 1 second. The input for extracting appr, place, and clip features is 1 frame, and for extracting the action features is a sequence of 16 frames. Audio features are extracted every 1 second, and the audio model's input is a window of 10 seconds. Text features are extracted for each subtitle timed text and each sentence of the synopsis. For each shot, we assign the features whose input has overlap with the shot time interval. E.g., audio is split into 10\textit{s} (seconds) windows with an overlap of 9\textit{s} (stride =1\textit{s}). Then, we get the features for windows which overlap with each shot, or for subtitles, we get the features for the subtitle timed segments whose time interval overlaps with each shot time interval.

\section{Expectation Step}

For synchronization between synopsis sentences and movie shots, we use Eq.~\ref{eq:em} as the objective. This objective is solved in an alternative manner, where we estimate the target variable $W$ via fixed parameters in the first step (E-step), and update the parameters while the target variable is known (M-step). 
\begin{equation}
\label{eq:em}
    \begin{split}
        \underset{W,\theta}{\mathrm{max}} & \sum_{i,j} w_{ij}   F(.;\theta) - \lambda \sum_{i, j} | w_{i,j} |
        \\
        & s.t.\ 0\leq w_{ij} \leq 1
    \end{split}
\end{equation}
Let $M_{sim}$ with dimension $L_{sh} \times L_{syn}$ be the similarity matrix with $m_{ij}$ entries  representing the similarity between the $i$-th shot and $j$-th synopsis sentence for one sample. Assuming $F(.;\theta) = M_{sim}$, there is a closed form solution for Eq.~\ref{eq:em} in the expectation step. We also use~\cite{mirza2021alignarr} to reduce the search space during optimization to only the pairs which are inside a diagonal boundary, i.e., all $m_{ij}$ outside the diagonal boundary are ignored. Eq.~\ref{eq:solution_em} shows the closed form solution for the expectation step. Following~\cite{mirza2021alignarr}, $\xi$ is set to $0.3$.
\begin{equation}
\scriptsize
\label{eq:solution_em}
    w_{ij} = 
    \begin{cases}
    1 & \text{if $m_{ij} \geq \lambda \ \&\  j < i\frac{L_{syn}}{L_{sh}} + \xi L_{syn} \ \&\  i < j\frac{L_{sh}}{L_{syn}}  +\xi L_{sh}$}\\
    0 & \text{if $m_{ij} < \lambda \ ||\  j \geq i\frac{L_{syn}}{L_{sh}} + \xi L_{syn} \ ||\  i \geq j\frac{L_{sh}}{L_{syn}}  +\xi L_{sh}$}\\
    \end{cases}
\end{equation}

\noindent\textbf{Proof:}
Due to non-negative constraint $0\leq w_{ij} \leq 1$, the objective function in Eq.~\ref{eq:em} is reduced to Eq.~\ref{eq:s2}. 
\begin{equation}
\label{eq:s2}
    \begin{split}
        &\underset{W}{\mathrm{max}} \sum_{i,j} w_{ij}   (m_{ij} - \lambda)  
        \\
        & s.t.\ 0\leq w_{ij} \leq 1
    \end{split}
\end{equation}

This objective has a closed form solution, shown in Eq.~\ref{eq:closed_sol_em}. Combining this solution with the diagonal constraint from~\cite{mirza2021alignarr} the final solution boils down to Eq.~\ref{eq:solution_em}.
\begin{equation}
\small
\label{eq:closed_sol_em}
    w^*_{ij} = 
    \begin{cases}
    1 & \text{if $m_{ij} \geq \lambda$}\\
    0 & \text{if $m_{ij} < \lambda$}\\
    \end{cases}
\end{equation}

\section{Knowledge Transfer}
Knowledge distillation is used to transfer the  knowledge available for the training samples on synopsis. 
More concretely, the soft similarity scores for each shot are normalized using Eq.~\ref{eq:attn}, and the shot level probability scores are derived using Eq.~\ref{eq:kdlabs} for $i$-th shot and $n$-th TP.
Let $Y$
be the predicted logits from the shot model ($\in \mathbb{R}^{L_{sh} \times N_{tp}}$, with entries $y_{in}$ for the $i$-th shot and $n$-th TP)
, where a softmax function derives the probability predictions per shot and TP (Eq.~\ref{eq:softmaxoutput}).
Given predicted probabilities from the model on the shot level $O$ (
$\in \mathbb{R}^{L_{sh} \times N_{tp}}$, with entries $o_{in}$ for the $i$-th shot and $n$-th TP), Kullback–Leibler divergence loss between $O{.n}$ and $P{.n}$ for each of the turning points (i.e., each $n$) is minimized during training (Eq.~\ref{eq:kdloss}).

\begin{equation}
\label{eq:attn}
    a_{ij}=\frac{exp(u_iv_j/\tau)}{\sum_{k}^{L_{syn}}exp(u_iv_k/\tau)} 
\end{equation}

\begin{equation}
\label{eq:kdlabs}
    \begin{split}
    &p_{in} =\frac{exp\left(\sum_{k=1}^{L_{syn}}a_{ik}q_{kn}\right)}{\sum_{j=1}^{L_{sh}}exp\left(\sum_{k=1}^{L_{syn}} a_{jk} q_{kn} \right)} 
    \end{split}
\end{equation}

\begin{equation}
\label{eq:softmaxoutput}
    o_{in} = \frac{exp(y_{in})}{\sum_{k=1}^{L_{sh}}exp(y_{kn})}
\end{equation}

\begin{equation}
\label{eq:kdloss}
    \mathscr{L}_{kd} = \sum_{n=1}^{N_{tp}}\mathrm{KL} \left(O_{.n} || P_{.n}\right)
\end{equation}


\begin{table*}[h]
\small	
\centering
    \subfloat[Effectiveness of Alignment Positional Encoding.]
    {
    \scalebox{0.8}{
        \begin{tabularx}{0.65\textwidth}{lccccc}
        \toprule
        & \multicolumn{2}{c}{\textit{\textbf{Scene Seg.}}} & \multicolumn{3}{c}{\textit{\textbf{Act Segmentation}}}   \\ 
		\cmidrule(lr){2-3} \cmidrule(lr){4-6} 
        {\textit{case}}  &  {\textit{AP}} &  {\textit{F1}} & {\textit{TA}}&  {\textit{PA}} & {\textit{D}}\\
            \midrule
              w/o align. PE          & 57.77 & 54.75 & 5.29 (4.74) & 7.37 (0.75) & 31.04 (1.04)\\
              w/o align. PE(50)       &  -   & -  & 7.31 (1.13) & 10.26 (1.19) & 15.75 (1.17)\\
              w. align. PE             & 58.59 &55.29 & 13.93 (2.21)  & 20.72 (2.28) & 9.19 (0.58)\\
              
        \bottomrule	
        \end{tabularx}
        }
        \label{tab:norm_PE}
    }\hfill
    \subfloat[Effectiveness of Normalized Positional Encoding in bottleneck tokens.]
    {
    \scalebox{0.8}{
        \begin{tabularx}{0.65\textwidth}{llccccc}
        \toprule
        && \multicolumn{2}{c}{\textit{\textbf{Scene Seg.}}} & \multicolumn{3}{c}{\textit{\textbf{Act Segmentation}}}   \\ 
		\cmidrule(lr){3-4} \cmidrule(lr){5-7} 
        {\textit{case}}  & {\textit{modality}}&  {\textit{AP}} & {\textit{F1}}&  {\textit{TA}}&  {\textit{PA}} & {\textit{D}}\\
            \midrule
            w/o align.PE   & V        & 58.31  & 54.83& 13.60 (1.59) & 20.53 (2.27) & 9.47 (1.23)\\
              w. align.PE    & V        & 58.59 &55.29 &13.93 (2.21)  & 20.72 (2.28) & 9.19 (0.58)\\
            w/o align.PE   & V + T        & - & - & 13.01 (1.43) & 20.13 (2.33) & 9.56 (0.92)\\
            w. align.PE    & V + T    & - & -  & 14.63 (1.10) &   21.78 (1.22) &  8.96 (0.65)\\
        \bottomrule	
        \end{tabularx}
        }
        \label{tab:norm_PEFT}
    }\hfill
    \subfloat[Multi-modal fusion strategies.]
    {
    \scalebox{0.8}{
        \begin{tabularx}{0.65\textwidth}{lccccc}
        \toprule
        & \multicolumn{2}{c}{\textit{\textbf{Scene Seg.}}} & \multicolumn{3}{c}{\textit{\textbf{Act Segmentation}}}   \\ 
		\cmidrule(lr){2-3} \cmidrule(lr){4-6} 
        {\textit{MM. integ. type}}  &  {\textit{AP}}&  {\textit{F1}} & {\textit{TA}}&  {\textit{PA}} & {\textit{D}}\\
            \midrule
              LateFusion& 58.24 & 48.27& 12.57 (1.71) & 19.21 (2.34) & 10.00 (1.22) \\
              Bottleneck             & 58.59&55.29 & 13.93 (2.21)  & 20.72 (2.28) & 9.19 (0.58)\\
        \bottomrule	
        \end{tabularx}
        }
        \label{tab:MM_fusion}
    }
    \hfill
    \subfloat[Impact from input modalities on scene segmentation.]
    {
    \scalebox{0.9}{
        \begin{tabularx}{0.23\textwidth}{lcc}
        \toprule
        {\textit{change}}  &  {\textit{AP}}&  {\textit{F1}} \\
            \midrule
              -clip                   & 58.09 &53.95  \\
              -place            & 57.51 & 54.83\\
              -bassl            & 51.88 & 43.04\\
              -bassl(50)            & 53.80 & 43.67\\
              -clip-place            & 57.92 & 50.71 \\
            - & 58.59  &  55.29 \\
            & &    \\
            & &    \\
        \bottomrule	
        \end{tabularx}
        }
        \label{tab:modality_scene}
    }
    \subfloat[Impact from input modalities on act segmentation.]
    {
    \scalebox{0.9}{
        \begin{tabularx}{0.48\textwidth}{lccc}
        \toprule
        {\textit{change}}  &  {\textit{TA}} & {\textit{PA}} & {\textit{D}} \\
            \midrule
              -clip                   & 6.09 (0.86) &  10.66 (1.48) & 21.81 (4.30)\\
              -clip(20)                   & 11.99 (1.73) &  17.89 (3.04)& 10.32 (1.68)\\
              -place            & 13.57 (2.78) & 19.87 (4.00)& 9.22 (1.38) \\
              -action            & 13.31 (1.98) & 20.20 (2.90) & 10.38 (2.03) \\
              -appr            & 13.42 (2.34) & 20.59 (3.40) & 8.85 (0.90)\\
              - & 13.93 (2.21)  & 20.72 (2.28) & 9.19 (0.58)\\
              +subtitle  & 13.93 (2.21)  & 20.72 (2.28) & 9.19 (0.58)\\
              +subtitle+audio & 14.19 (1.13) &   22.10 (1.46) &  9.68 (1.06)\\
        \bottomrule	
        \end{tabularx}
        }
        \label{tab:modality_act}
    }\hfill
    \subfloat[Impact of Synchronization with multimodal video features.]
    {
    \centering
    \scalebox{1}{
        \begin{tabularx}{0.61\textwidth}{llccc}
        \toprule
        && \multicolumn{3}{c}{\textit{\textbf{Act Segmentation}}}   \\ 
		\cmidrule(lr){3-5} 
        {\textit{synopsis synch. by}}  &{\textit{M for synch.}}  & {\textit{TA}}&  {\textit{PA}} & {\textit{D}}\\
            \midrule
               ~\cite{papalampidi2019movie}  & T                   &  10.51 (0.72) & 14.54 (1.09) & 8.98 (0.25) \\
               MEGA & V         & 13.93 (2.21)  & 20.72 (2.28) & 9.19 (0.58) \\
               MEGA  & V + T       & 14.63 (1.10) &   21.78 (1.22) &  8.96 (0.65) \\
        \bottomrule	
        \end{tabularx}
        }
        \label{tab:align}
    }\hfill 
    \subfloat[Impact from feature set and model size on scene seg. SPS denotes $\#$ of samples per second.]
    {
    \scalebox{0.9}{
        \begin{tabularx}{0.6\textwidth}{llcccc}
        \toprule
        {\textit{Approach}}  & {\textit{Feature Set Pretrained on}}  &  {\textit{AP}} &  {\textit{F1}}  &  {\textit{Params}}   &  {\textit{SPS}}\\
            \midrule
            BaSSL~\cite{mun2022boundary}         &    Movienet      & 57.4 & 47.02 & 15.77M & 6244.99 \\
              LGSS~\cite{rao2020local}         &   M+P+I        & 52.93 & 48.75 &66.16M & 206.36 \\
              MEGA & M+P+I & 58.59 &55.30 &  67.57M & 1736.13 \\
        \bottomrule	
        \end{tabularx}
        }
        \label{tab:feat_size_scene}
    }\hfill
    \subfloat[Impact from feature set and model size on act seg. Set1 includes Visual (appr), Audio (YAMNet), Textual (script-USE). Set2 has Visual (appr,clip,action,place), Audio (audio), Textual (text from subtitle).]
    {
    \scalebox{0.9}{
        \begin{tabularx}{0.72\textwidth}{llccccc}
        \toprule
        {\textit{Approach}} & {\textit{Feature Set}}   &  {\textit{TA}} & {\textit{PA}} & {\textit{D}} & {\textit{Params}} & {\textit{SPS}} \\
            \midrule
            GRAPHTP~\cite{papalampidi2021movie}             &  Set1~\cite{papalampidi2021movie}    & 9.12 &  12.63 & 9.77 & 0.745M & 25.40 \\
            GRAPHTP~\cite{papalampidi2021movie}             &  Set2    & 4.72 &  7.37 & 9.69 & 6.78M & 14.36 \\
            MEGA             &  Set2    & 14.19 (1.13) &   22.10 (1.46) &  9.68 (1.06) & 6.78M & 18.24 \\
        \bottomrule	
        \end{tabularx}
        }
        \label{tab:feat_size_act}
    }
    \vspace{-0.2cm}
    \caption{Ablation studies on MEGA components.  Values within parentheses are standard deviations for multiple runs.}
    \vspace{-0.2cm}
    \label{tab:ablation}
\end{table*}

\section{Experiments}
\subsection{Additional Results}

Unless otherwise specified, this section includes further statistics and metrics for the same experiments which are provided in the paper. 

Tab.~\ref{tab:ablation} includes the $F1$ score for Scene Segmentation model in all the ablation experiments. The ablation experiments show a similar trend across $AP$ and $F1$ score.  Tab.~\ref{tab:modality_scene} shows an extra experiment with ablation of bassl features (i.e., -bassl(50)). In this experiment, we further inspect this ablation by continuing the training for this model for 50 epochs, which still shows a significant difference with the model with all the modalities included, demonstrating the effectiveness of multimodal fusion in MEGA. 

Additionally, given that all experiments for Act Segmentation with MEGA are repeated 8 times, and the performance metrics are averaged, the standard deviations (STD) of all such experiments  for comparison with previous SoTA and in ablation experiments are provided in parentheses in Tabs.~\ref{tab:sotatpd} and~\ref{tab:ablation}, respectively. Considering the STD values, results still demonstrate that MEGA outperforms previous SoTA on act segmentation, and the ablations of various components in MEGA worsen the performance, showing the importance of each of those components.
Additionally, Tab.~\ref{tab:norm_PE} includes an extra experiment (i.e., w/o align. PE(50)), where we further investigate the ablation of this module for act segmentation by training the model without the normalized positional encoding for longer time (50 epochs). By further training, the performance gap reduces but still remains to be significant, which indicates that the proposed normalized positional encoding not only helps the model converge faster but also is a necessary component. Furthermore, Tab.~\ref{tab:modality_act} shows an extra experiment (-clip(20)), where we further trained the model for 20 epochs. The results are improved but there is still a significant gap with the model which uses clip, showing that the importance of multimodal fusion in MEGA.


\begin{table*}[h]
\centering
\scalebox{1}{
    \centering
    \begin{tabular}{lllccc}
    \hline
    \multirow{2}{*}{\textbf{\textit{Approach}}} & \multirow{2}{*}{\textbf{\textit{Modality}}} &\textbf{\textit{Modality}} & \textbf{\textit{TA}}  & \textbf{\textit{PA}} & \textbf{\textit{D}}  \\
          & & \textbf{\textit{for synch.}}  & \textbf{\textit{[\%]}} & \textbf{\textit{[\%]}} & \textbf{\textit{[\%]}} \\
        \hline 
    Random (Even. distribution)~\cite{papalampidi2021movie}& -  & T*&4.82 &6.95& 12.35 \\
    Theory~\cite{hauge2017storytelling,papalampidi2019movie}& - & T*& 4.41 &6.32  & 11.03\\
    Distribution position~\cite{papalampidi2021movie}& - & T*&5.59&  7.37 & 10.74\\
    \hline
    \multicolumn{4}{l}{\textbf{Single modality input}} \\
 
    TEXTRANK~\cite{mihalcea2004textrank}& T & T*& 6.18 & 10.00 & 17.77 \\
    SCENESUM~\cite{gorinski2015movie}& T  & T*& 4.41 & 7.89&  16.86\\
    TAM~\cite{papalampidi2020screenplay}& T & T*& 7.94 & 9.47 & 9.42\\
    GRAPHTP~\cite{papalampidi2021movie} & T & T*& 6.76 & 10.00 & 9.62 \\  
	\rowcolor{Gray}
    MEGA* & V  & T*& 10.51 (0.72) & 14.54 (1.09) & 8.98 (0.25)\\
	\rowcolor{Gray}
    MEGA & V & V& 13.93 (2.21)  & 20.72 (2.28) & 9.19 (0.58) \\
    \hline
    \multicolumn{4}{l}{\textbf{Multi-modality input}} \\
    TEXTRANK~\cite{mihalcea2004textrank}& T+A+V & T*& 6.18 & 10.00 & 18.90\\
    SCENESUM~\cite{gorinski2015movie}& T+A+V& T* &6.76  & 11.05 &18.93 \\
    TAM~\cite{papalampidi2020screenplay}& T+A+V & T*& 7.36 & 10.00 & 10.01\\
    GRAPHTP~\cite{papalampidi2021movie} & T+A+V& T*& 9.12 & 12.63 & 9.77 \\
	\rowcolor{Gray}
    MEGA* & T+V  & T*& 11.14 (1.77) & 15.20 (2.33) & \textbf{8.96  (0.35)} \\
	\rowcolor{Gray}
    MEGA & T+V & T+V&  \textbf{14.63 (1.10)} &   21.78 (1.22)  &  \textbf{8.96 (0.65)} \\
	\rowcolor{Gray}
    MEGA* & T+A+V & T* & 10.00 (0.98) &  14.08 (1.56)&  8.96 (0.39)\\
	\rowcolor{Gray}
    MEGA & T+A+V & T+A+V&  14.19 (1.13) &   \textbf{22.10 (1.46)} &  9.68 (1.06)\\
    \hline
    \end{tabular}
}
    \caption{TP identification: comparison with SoTA. MEGA* denotes the MEGA using the same synchronization as in~\cite{papalampidi2019movie} for fair comparison. T*,V,T,A denote Textual-screenplay, Visual, Textual-subtitle and Acoustic features respectively. Values within parentheses are standard deviations for multiple runs.}
     \label{tab:sotatpd}
\end{table*}

\subsection{Hyperparameter Search in Align. PE. for $L_n$} 

\begin{figure}
    \centering
    \includegraphics[scale=0.60]{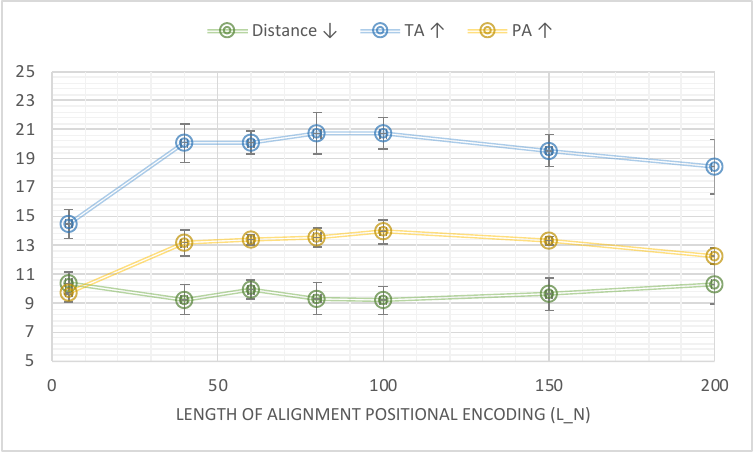}
    \caption{Evaluation of act segmentation in terms of Distance, TA, and PA metrics by changing $L_n$.}
    \vspace{-.5cm}
    \label{fig:act_ln}
\end{figure}

We demonstrate the performance of act segmentation model with different values for $L_n$ (i.e., length of Align. PE.) in Fig.~\ref{fig:act_ln}. The experiment results reveal that while MEGA performs best at $L_n=100$ for act segmentation, it performs robustly across a range of $L_n$ values (specifically, within the range of $50$ to $150$). Align. PE. is designed to provide  video-level coarse progress information as a complementary signal next to regular PE. If $L_n$ is very small, the granularity of the align. PE. becomes too coarse, whereas if $L_n$ is too large, it becomes overly detailed as regular PE; which explain the performance decline we observe in Fig.~\ref{fig:act_ln} at small and big $L_n$ values. Aside from better performance, using $L_n \ll L$, where $L$ is the input length, results in better efficiency. During multimodal fusion, the number of fusion tokens is equal to $L_n$, hence the memory consumption during multimodal fusion attention calculation is $(L_n \times m + L) ^ 2$, where $m$ is the number of modalities. Our approach enables aligned multimodal alignment during fusion that scales efficiently with respect to the number of modalities in terms of memory consumption.

\subsection{Visualization}
\subsubsection{Feature Importance}

To further look into how MEGA is integrating different modalities to make a prediction, we calculated the GradCAM~\cite{selvaraju2017grad} for the output features from the fusion module. 
More concretely, the derivative of the outputs (the maximum prediction logit of the two dimensional output for \emph{scene segmentation} and the max predicted shot for \emph{act segmentation}) with respect to the final FC layer parameters are calculated. 
These values are then multiplied with activation scores coming out of each modality fusion module, summed across the channel dimension and undergone a ReLU non-linearity function. 
The value scores are then normalized across all modalities, such that their summation is 1. This helps to visualize the effect of each of the modalities in the prediction from the model.

\begin{figure*}[b]
    \centering
    \includegraphics[scale=0.19]{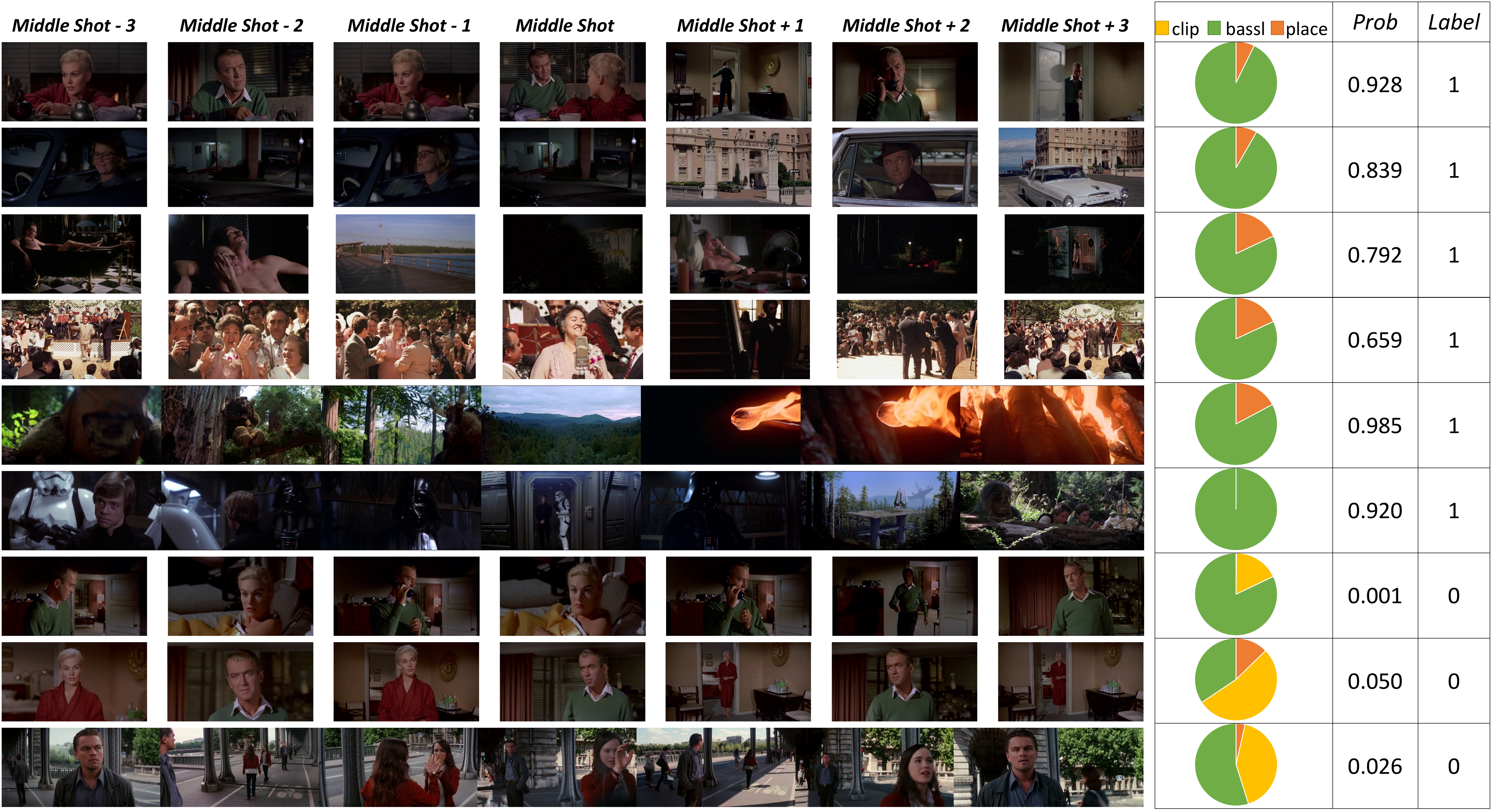}
    \caption{GradCAM values shown in pie charts for 9 different predictions on scene segmentation on the test set, along with the probability prediction score for the middle shot being the end of a scene and its groundtruth label.}
    \label{fig:vis_scene}
\end{figure*}

\begin{figure*}
    \centering
    \includegraphics[scale=0.18]{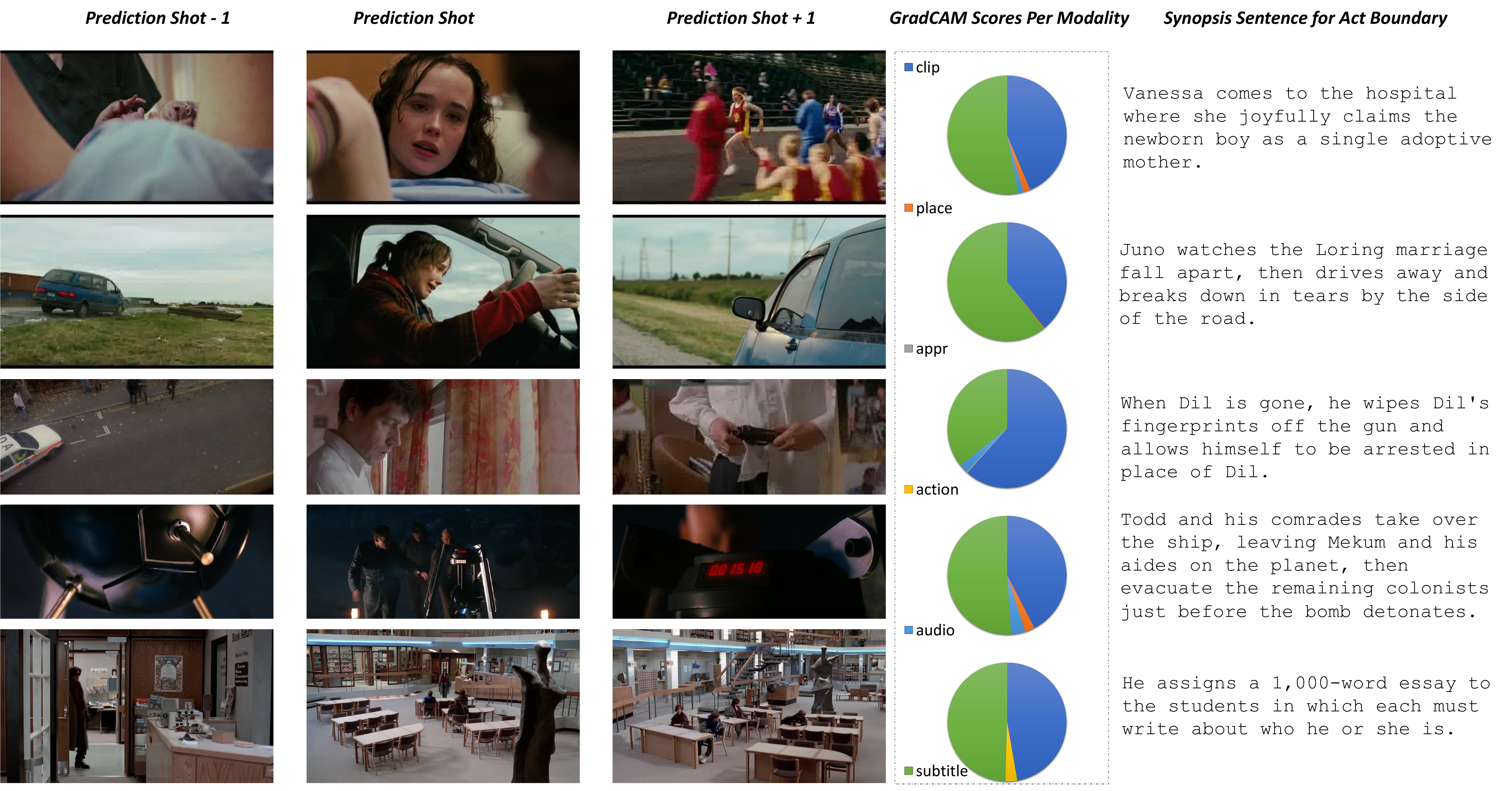}
    \caption{GradCAM for 5 different predicted turning points on the test set, along with their synopsis annotated sentence for that turning point.}
    \label{fig:vis_tp}
\end{figure*}

For scene segmentation, Fig.~\ref{fig:vis_scene} demonstrates the results. The results in this figure are aligned with Tab.~\ref{tab:modality_scene} showing the order of importance for the modalities are bassl, place and clip. 
For act segmentation, Fig. \ref{fig:vis_tp} demonstrates the results for the 5 predicted act boundaries. 
The demonstrations are aligned with the ablation experiments in Tab.~\ref{tab:modality_act}, showing the highest contributions are from the clip and subtitle modalities.

\subsubsection{Expected Synchronization Matrix}
\begin{figure*}
    \centering
    \includegraphics[scale=0.5]{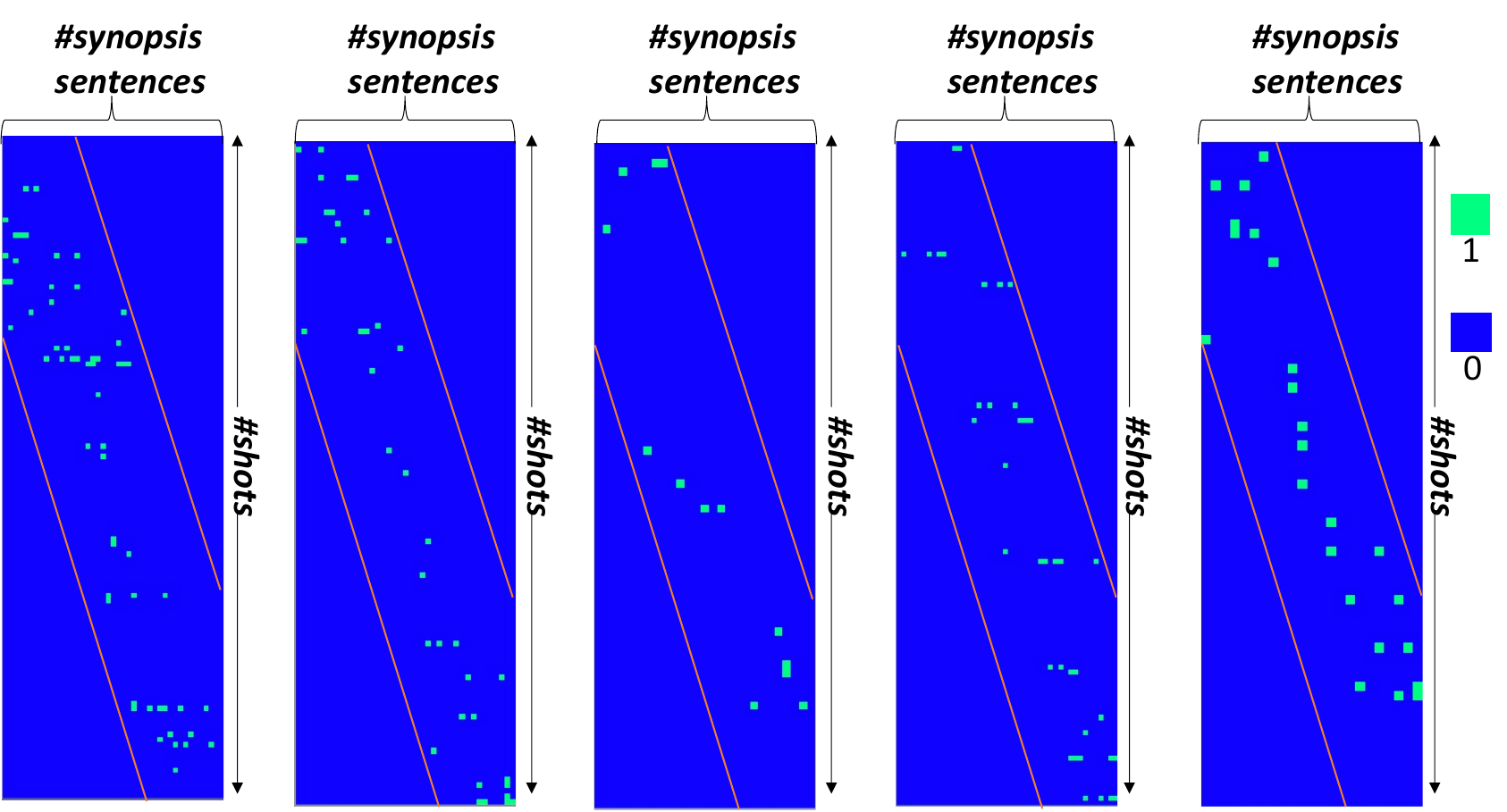}
    \caption{The expected value of synchronization matrix for 5 different samples on the test set. The actual matrices had to be resized for visualization (ratio of height/width is set to 3).}
    \label{fig:exp}
\end{figure*}

Fig.~\ref{fig:exp} shows the expected value of synchronization matrix during optimization on different samples from TRIPOD test set. The results demonstrate that the synchronization matrix synchronizes the synopsis sentences and shots more along the diagonal line, which is expected. 

\subsubsection{Attention Scores}
\begin{figure*}
    \centering
    \includegraphics[scale=0.44]{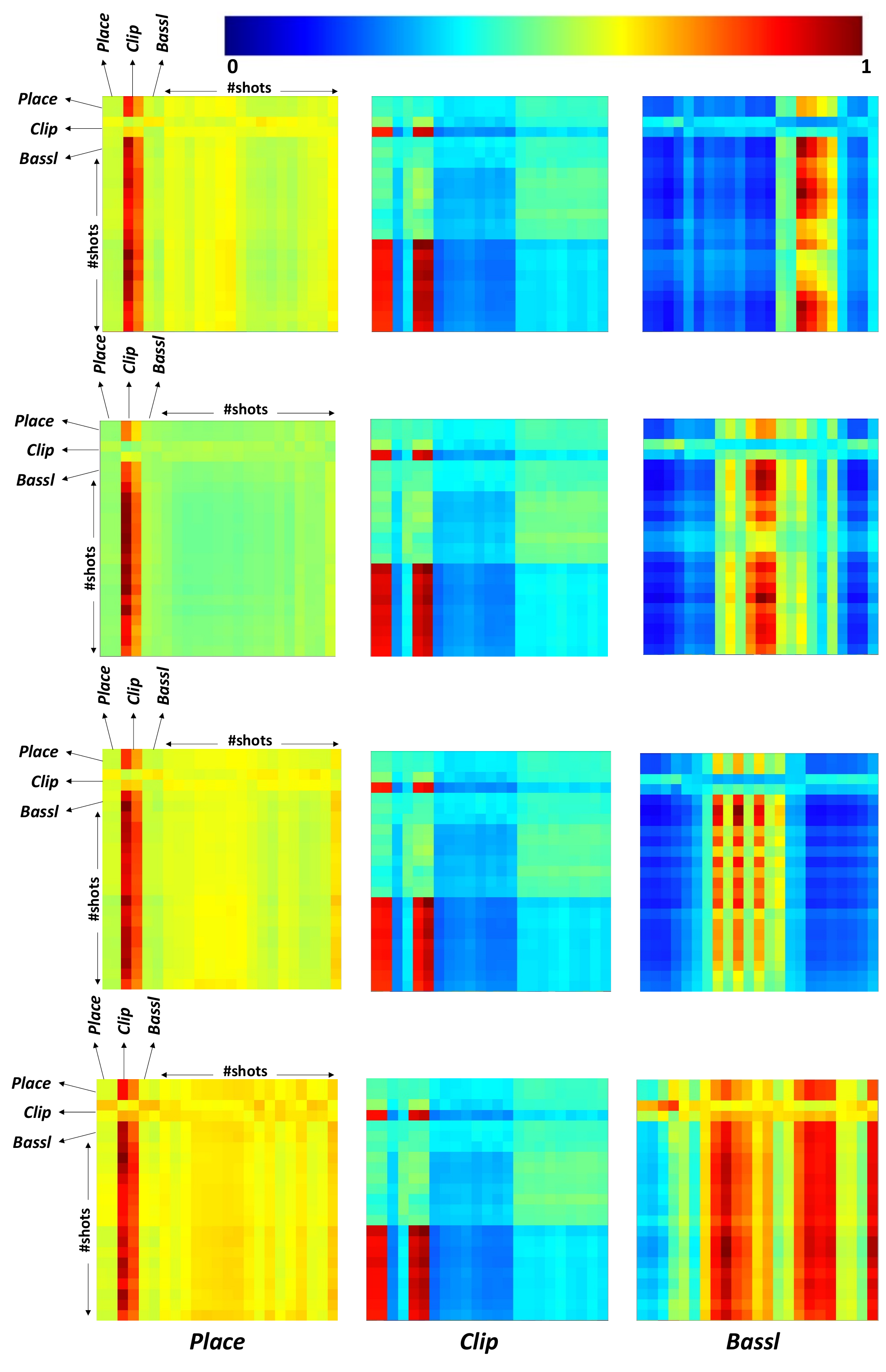}
    \caption{Attention scores derived from the fusion transformer encoder on Scene Segmentation model.}
    \label{fig:attn_movient}
\end{figure*}

\begin{figure*}
    \centering
    \includegraphics[scale=0.3]{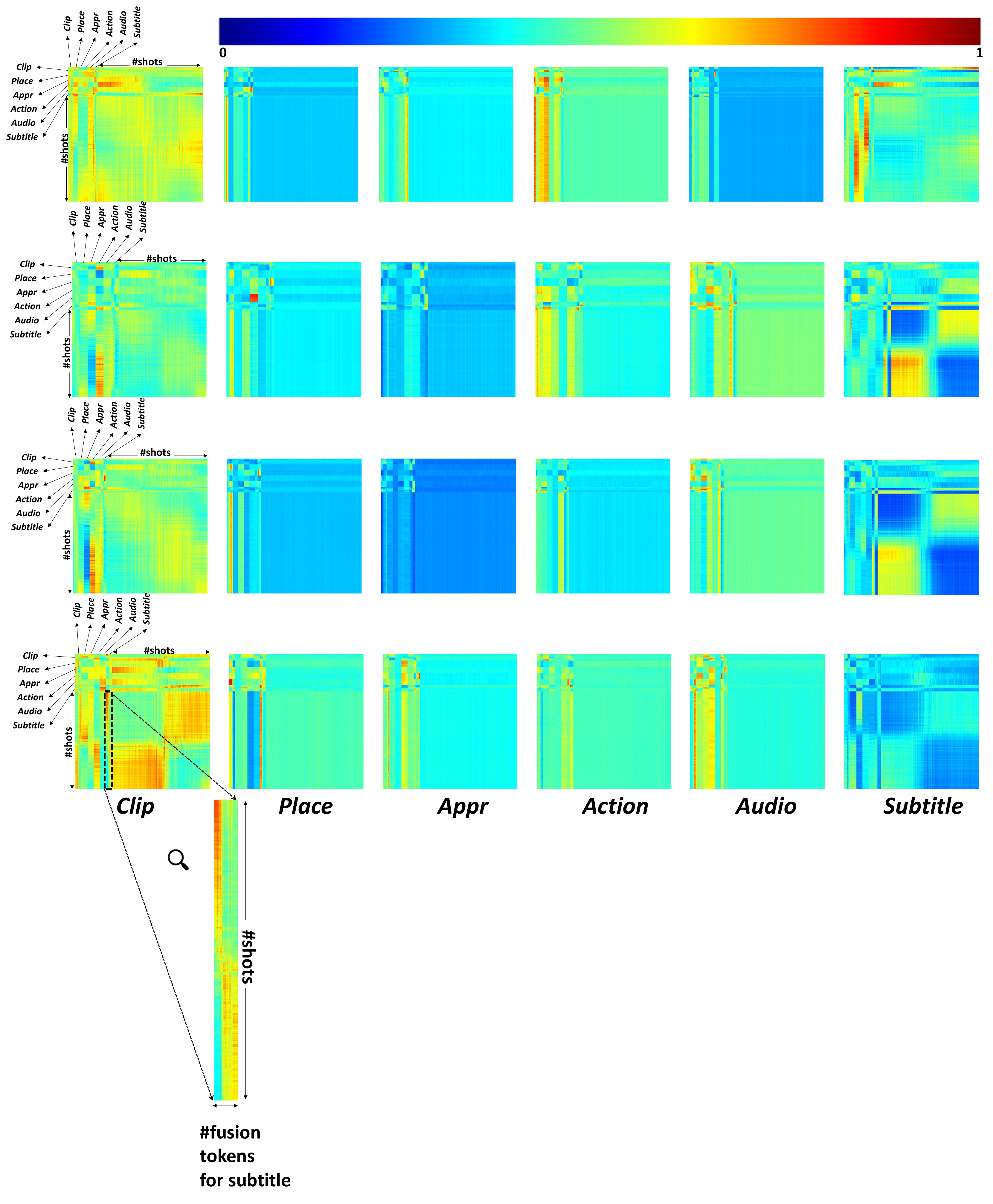}
    \caption{Attention scores derived from the fusion transformer encoder on Act Segmentation model.}
    \label{fig:attn_tripod}
\end{figure*}

Figs.~\ref{fig:attn_movient} and~\ref{fig:attn_tripod} demonstrate the attention scores for the fusion modules across the Scene Segmentation and Act Segmentation models on several test samples. For better visualization, all scores within one image are normalized by their max value within that image. These figures clearly demonstrate that the model is fusing different modalities flexibly for different time units (i.e., shots). And, while for some of the modalities the fusion patterns remains more similar across different time units, for some there is a clear change in pattern across time (e.g., see the zoomed area in the last row of Clip attention, which shows MEGA's fusion tokens have the capability to preserve temporal information).

\clearpage

{\small
\bibliographystyle{ieee_fullname}
\bibliography{references}
}

\end{appendices}

\end{document}